\documentclass{article}

\usepackage{microtype}
\usepackage{graphicx}
\usepackage{subcaption}
\usepackage{booktabs} % for professional tables
\usepackage{url}
\usepackage[utf8]{inputenc} % allow utf-8 input
\usepackage[T1]{fontenc}    % use 8-bit T1 fonts
\usepackage{amsfonts}       % blackboard math symbols
\usepackage{nicefrac}       % compact symbols for 1/2, etc.
\usepackage{microtype}      % microtypography
\usepackage{xcolor}         % colors
\usepackage{multirow}
\usepackage{amsmath}
\usepackage{subcaption} % For subfigures
\usepackage{colortbl}
\usepackage{amssymb}
\usepackage{bbding}
\usepackage{pifont}
\usepackage[percent]{overpic}  % Add this to preamble
\usepackage{makecell}

\newcommand{\secondbest}[1]{\underline{#1}}
\newcommand{\best}[1]{\textbf{#1}}
\DeclareSymbolFont{extraup}{U}{zavm}{m}{n}
\DeclareMathSymbol{\newcrossmark}{\mathalpha}{extraup}{129}%uni2717

\usepackage{hyperref}

\usepackage[preprint]{icml2026}

\usepackage[capitalize,noabbrev]{cleveref}

\usepackage[textsize=tiny]{todonotes}

\icmltitlerunning{DiSa: Saliency-Aware Foreground-Background Disentangled Framework for Open-Vocabulary Semantic Segmentation}

\begin{document}

\twocolumn[
\icmltitle{DiSa: Saliency-Aware Foreground-Background Disentangled Framework for Open-Vocabulary Semantic Segmentation}

\begin{icmlauthorlist}
\icmlauthor{Zhen Yao}{xxx}
\icmlauthor{Xin Li}{yyy}
\icmlauthor{Taotao Jing}{yyy}
\icmlauthor{Shuai Zhang}{yyy}
\icmlauthor{Mooi Choo Chuah}{xxx}
\end{icmlauthorlist}

\icmlaffiliation{xxx}{Department of Computer Science and Engineering, Lehigh University}
\icmlaffiliation{yyy}{Qualcomm AI Research}

\icmlcorrespondingauthor{Zhen Yao}{zhy321@lehigh.edu}

\icmlkeywords{Foreground-Background Disentanglement, Vision-Language Model, Open-Vocabulary Semantic Segmentation}
\vskip 0.3in
]

\printAffiliationsAndNotice{}  % no special notice 
% \maketitle
\begin{abstract}
Open-vocabulary semantic segmentation aims to assign labels to every pixel in an image based on text labels. Existing approaches typically utilize Vision-Language Models (VLMs), such as CLIP, for dense prediction. However, VLMs, pre-trained on image-text pairs, are biased toward salient, object-centric regions and exhibit two critical limitations when adapted to segmentation: (i) \textbf{\textit{Foreground Bias}}, which tends to ignore background regions, and (ii) \textbf{\textit{Limited Spatial Localization}}, resulting in blurred object boundaries. To address these limitations, we introduce \textbf{DiSa}, a novel saliency-aware foreground-background disentangled framework. By explicitly incorporating saliency cues in our designed Saliency-aware Disentanglement Module (SDM), DiSa separately models foreground and background ensemble features in a divide-and-conquer manner. Additionally, we propose a Hierarchical Refinement Module (HRM) that leverages pixel-wise spatial contexts and enables channel-wise feature refinement through multi-level updates. Extensive experiments on six benchmarks demonstrate that DiSa consistently outperforms state-of-the-art methods.
\end{abstract}    
\vspace{-1.8\baselineskip}
\section{Introduction}  \label{sec:intro}
\vspace{-0.2\baselineskip}
Open-vocabulary semantic segmentation aims to label each pixel with an unlimited range of categories that extend beyond a pre-defined closed set, based on text labels. To this end, vision-language models (VLMs), e.g., CLIP \citep{radford2021learning} and ALIGN \citep{jia2021scaling}, have been widely explored, as they exhibit powerful zero-shot recognition capabilities via large-scale training on image-text pairs. \par

\begin{figure}[!t]
\centering
\begin{overpic}[width=0.88\linewidth]{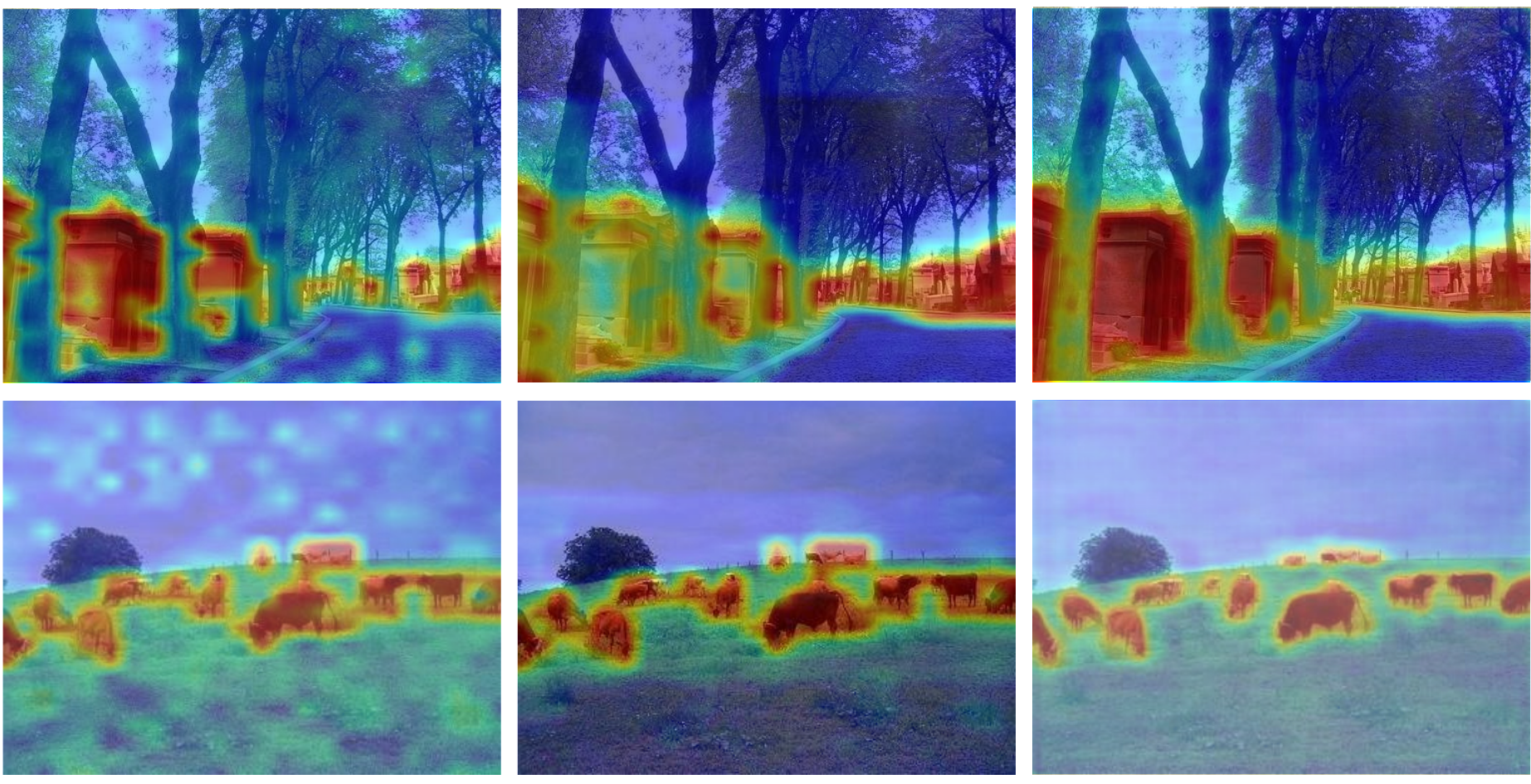}
  \put(9, -3.6){\small (a) CLIP}       % left column caption
  \put(39, -3.6){\small (b) CAT-Seg}  % middle column
  \put(75, -3.6){\small (c) DiSa}  % right column
\end{overpic}
\vspace{0.5\baselineskip}
\caption{\textbf{Visualization of correlation maps.} VLMs face \textbf{\textit{Foreground Bias}} and \textbf{\textit{Limited Spatial Localization}} challenges. Our proposed DiSa effectively alleviates these challenges. Row 1 indicates class “building”, and Row 2 indicates class “animal”. }
\label{fig:motivation}
\vspace{-1.4\baselineskip}
\end{figure}

Despite these advances, VLMs pre-trained on image-text pairs face 2 critical limitations when adapted to dense prediction tasks: (1) \textbf{\textit{Foreground Bias}}: VLMs tend to overemphasize salient, foreground regions while neglecting background context, leading to misclassification of background regions \citep{li2024densevlm}. This bias stems from the object-centric nature of pre-training data, where captions predominantly describe salient, foreground instances. This results in a fundamental misalignment between the foreground-centric bias of VLMs and the pixel-level precision of segmentation, which requires holistic scene understanding and accurate recognition of non-salient background regions. As shown in row 1 of Fig. \ref{fig:motivation}, VLMs pay little attention to non-salient, background buildings. (2) \textbf{\textit{Limited Spatial Localization}}: VLMs demonstrate limited capacity of fine-grained spatial reasoning required for segmentation predictions. Due to insufficient dense supervision during pre-training, these models struggle to capture precise object boundaries and reconstruct local structural details (as shown in row 2 of Fig. \ref{fig:motivation}). This poses challenges in distinguishing visually similar or spatially overlapping categories \citep{lee2025effective}, particularly for small objects and background regions that require nuanced spatial reasoning for accurate segmentation \citep{zhou2022extract, zhong2022regionclip}. \par

To address these limitations, we propose foreground-background disentanglement mechanisms to tackle various category roles across different visual contexts. Our design is motivated by the observation that most categories exhibit context-dependent roles, e.g., cars or furniture may appear as either foreground instances or background context depending on scene composition. This contextual distinction highlights the importance of adaptive representation learning that captures both fine-grained localization for foreground instances and the semantic coherence for background regions. While existing approaches explore token-level or class-level disentanglement, they either fail to preserve intra-class relationships or assume rigid taxonomies, leading to sub-optimal performance. To overcome these limitations, we leverage saliency cues for adaptive foreground-background inter-token decomposition for each category. Unlike prior works that merely employ saliency for computational efficiency \citep{choi2024salience, luo2024emergent}, our method explicitly leverages saliency to address the aforementioned \textbf{\textit{Foreground Bias}} challenge. Specifically, we leverage saliency maps derived from text-image cross-attention to effectively partition per-class visual embeddings into foreground (salient, object-centric) and background (contextual, peripheral) regions based on their corresponding saliency scores. This enables ensemble feature learning via dual branches that capture domain-specific characteristics while preserving semantic coherence. \par

Building on the saliency-aware disentanglement, we propose a novel framework, DiSa, which explicitly separates foreground and background features. This decomposition enables our model to learn distinct and complementary representations, addressing the inherent imbalance in semantic granularity between foreground and background regions. In addition, we introduce a Hierarchical Refinement Module (HRM) that captures detailed spatial context and refines features via multi-level updates. Specifically, it consists of (1) Pixel-wise Refinement, which enhances spatial localization at the pixel level; (2) Category-wise Refinement, which captures channel-wise coherence for each class; and (3) Semantic-wise Refinement, which extracts semantic consistency within broader foreground/background groupings. \par

In summary, our contributions in this paper include:
\vspace{-1\baselineskip}
\begin{itemize}
\item We propose \textbf{DiSa}, a novel Saliency-aware Foreground-background Disentangled framework for OVSS. Our Saliency-aware Disentanglement Module (SDM) is the first to use explicit saliency cues for adaptive intra-class foreground-background decomposition, enabling context-dependent assignment. It facilitates semantic coherence especially for non-salient background regions, mitigating \textbf{\textit{Foreground Bias}}.
\item DiSa introduces a Hierarchical Refinement Module (HRM) that captures spatial context through Pixel-, Category-, and Semantic-wise Refinement. By incorporating spatial and channel-level context modeling, HRM alleviates the challenge of \textbf{\textit{Limited Spatial Localization}}, improving fine-grained boundary localization and spatial discrimination capabilities.
\item We conduct extensive evaluations across six large-scale open-vocabulary semantic segmentation benchmarks. DiSa consistently outperforms state-of-the-art methods, achieving significant performance gains that demonstrate its effectiveness and robustness.
\end{itemize}
\vspace{-1\baselineskip}
\section{Related Work}  \label{sec:related}
\vspace{-0.5\baselineskip}
\subsection{Open-vocabulary Semantic Segmentation}\label{subsec:ovss}
With the advance of VLMs, researchers have started to explore their powerful visual-text alignment capabilities to provide semantically rich and aligned multimodal representations in this task. SegCLIP \citep{liu2024open} integrates CLIP with Vision Transformers (ViT) \citep{dosovitskiy2020image} through a semantic group module that aggregates patches with learnable centers. It additionally introduces two auxiliary losses, one is a reconstruction loss for recovering the masked patches, and another is a superpixel-based KL divergence loss. CAT-Seg \citep{cho2024cat} estimates cost volumes from CLIP image-text similarities, followed by spatial and class aggregation to improve localization accuracy. The cost volumes serve as visual groundings for class-specific predictions. SCAN \citep{liu2024open} presents a Semantic-assisted Calibration Network to mitigate misalignment between visual contents and text semantics by calibrating the mask proposals and reducing domain bias in CLIP. ESC-Net \citep{lee2025effective} leverages CLIP-derived image-text correlations as pseudo-supervision for SAM, generating accurate predictions through the powerful segmentation capabilities of foundation models. \par

A parallel line of research investigates saliency for computational efficiency. SBAM \citep{choi2024salience} proposes an adaptive masking mechanism based on saliency-driven importance scores to enhance pre-training efficiency. PnP-OVSS \citep{luo2024emergent} introduces a token pruning strategy that constructs class-agnostic saliency maps by aggregating category-specific attention, pruning less discriminative tokens. However, these approaches merely focus on computational optimization rather than leveraging saliency information to address critical limitations in foreground-background disambiguation and spatial localization inherent in VLMs. \par

\vspace{-0.5\baselineskip}
\subsection{Foreground-background Disentanglement}\label{subsec:disen}
\vspace{-0.5\baselineskip}
Recent works identified \textbf{\textit{Foreground Bias}} \citep{li2024densevlm} as a fundamental limitation in VLMs, where pre-training on object-centric image-text pairs introduces systematic biases toward salient regions while neglecting holistic scene understanding. To address this issue, researchers explored foreground-background decomposition strategies for ensemble modeling of background regions. \par

\begin{figure*}[!t]
\centering
\includegraphics[width=\linewidth]{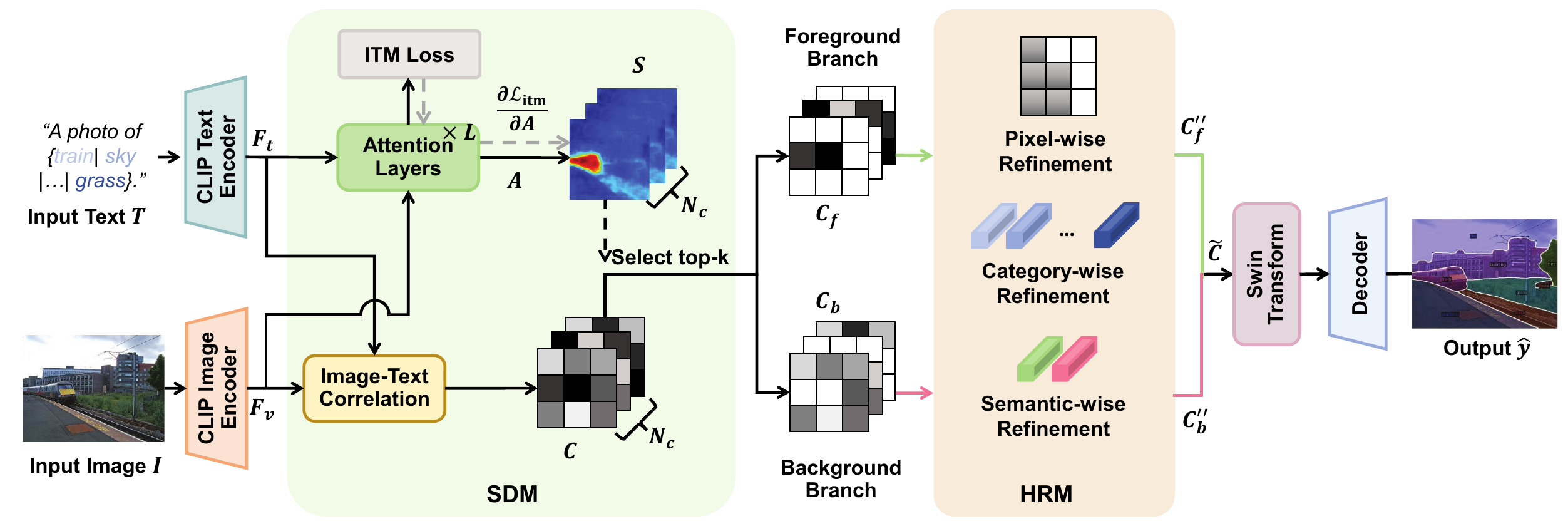}
\caption{\textbf{Overview of the  DiSa Framework.} DiSa consists of a Saliency-aware Disentanglement Module (SDM) and a Hierarchical Refinement Module (HRM), followed by an upsampling decoder.}
\vspace{-1.2\baselineskip}
\label{fig:model}
\end{figure*}

One research direction focuses on decomposing all visual embeddings into foreground and background regions at the token level. Panoptic SegFormer \citep{li2022panoptic} presents a query decoupling strategy to adaptively separate visual tokens into thing and stuff queries. OpenSeeD \citep{zhang2023simple} employs language guidance to select foreground queries, which subsequently interact with learnable background queries through decoupled cross-attention blocks. FOUND \citep{simeoni2023unsupervised} extracts high-confident “seed” tokens to generate coarse background masks through attention maps, enhancing fine-grained object localization. ClearCLIP \citep{lan2024clearclip} decomposes CLIP vision encoder outputs to attention output and residual connections, learning more robust object recognition. FOCUS \citep{you2025focus} leverages two pre-defined prompts to generate foreground/background masks and further calculates the contrastive loss for improved foreground localization. \par

Alternative approaches perform disentanglement at the class level, separating all classes into foreground and background taxonomies. DenseVLM \citep{li2024densevlm} designs a novel VLM that mitigates background imbalance by generating pseudo labels for unlabeled regions using frozen VLM and then applying separate alignment objectives for pre-defined foreground and background categories. Talk2DINO \citep{barsellotti2024talking} designs a Background Cleaning Procedure that re-weights class scores based on the self-attention maps, highlighting the foreground regions while suppressing background interference. LBP \citep{li2024learning} enhances background understanding for open-vocabulary object detection by learning background prompts from other images, effectively incorporating implicit background knowledge and achieving superior performance. \par

Despite these advances, existing disentanglement approaches suffer from several limitations. Token-level disentanglement fails to preserve intra-class relationships that are crucial for dense predictions. Class-level disentanglement assumes rigid foreground-background taxonomies, ignoring that instances of each class may appear as either foreground or background depending on scene composition. Furthermore, learnable disentanglement modules without any prior knowledge lack explicit guidance and are sub-optimal. In contrast to these approaches, our method leverages explicit saliency supervision to perform adaptive, class-specific foreground-background disentanglement, developing semantic coherence among ensemble representations while addressing the \textbf{\textit{Foreground Bias}} in VLMs. \par
\vspace{-0.5\baselineskip}
\section{Methodology}  \label{sec:method}
\vspace{-0.5\baselineskip}
\subsection{Motivation}
Our approach is motivated by the observation that the challenges of open-vocabulary semantic segmentation lie in the inherent asymmetry between foreground objects (salient, instance-centric elements) and background regions (contextual, peripheral environments), which are often entangled within shared feature spaces. To be more specific, in real-world images, the role of a category can vary based on scene composition (e.g., a train in focus vs. a train in the far background). This aligns with how humans parse visual scenes: we don’t always treat “train” in different scenes with equal attention — it depends on salience, size, occlusion, etc. \par

While existing methods treat all regions uniformly, we argue that the inherent differences between foreground and background semantics reveal the benefits of a more principled decomposition strategy. This allows both foreground and background features to learn specialized representations and learning objectives, aligning with the “Seek common ground while reserving differences” design ideology \citep{li2022panoptic}. Even for \textit{stuff} classes, the foreground refers to the most semantically informative or attribute-rich sub-regions. This behavior aligns with our above motivation: even within the same class, different regions may contribute differently to the textual concept. For example, in classes like wall or sky, the textured parts of a wall and the cloud structures in the sky provide relatively stronger visual cues, while others serve as contextual or peripheral background. \par

To this end, we propose a divide-and-conquer formulation that leverages saliency cues to structurally decompose the segmentation task into two complementary sub-problems: foreground object localization and background region understanding. This separation not only improves robustness to dynamic scene compositions but also enhances holistic scene understanding by improving disambiguation. \par

\vspace{-0.5\baselineskip}
\subsection{Architecture Overview}
\vspace{-0.5\baselineskip}
Fig. \ref{fig:model} provides an overview of our proposed framework, DiSa. Our model consists of 4 core components: a CLIP image encoder, a CLIP text encoder, a Saliency-aware Disentanglement Module (SDM) for foreground-background disentanglement, and a Hierarchical Refinement Module (HRM) for integrating multi-level fine-grained details. We follow existing works \citep{xian2019semantic, bucher2019zero} for the task design, e.g., the input and output formats. Given an input image $I$ and a set of text labels $T=\{T_i, i=1,2,...,{N_C}\}$, where $N_C$ is the number of all $C$ classes, we utilize CLIP as vision-language encoders to extract image $F_v \in \mathbb{R}^{H \times W \times D}$ and text embeddings $F_t \in \mathbb{R}^{N_C \times D}$, where $D$ is the dimension size. \par

Our proposed pipeline begins by processing image $F_v$ and text embeddings $F_t$ to extract cross-attention maps $A \in \mathbb{R}^{HW \times N_C}$ through SDM. These attention maps are then sharpened by Image-Text Matching (ITM) loss \citep{li2021align} gradients and the outputs are saliency maps $S_{1:N_C} \in \mathbb{R}^{H \times W \times N_C}$. Meanwhile, DiSa generates correlation maps $C_{1:N_C} \in \mathbb{R}^{H \times W \times N_C \times D}$ between image $F_v$ and text embeddings $F_t$ through cosine similarity and projection MLP layers. All correlation tokens from $C$ are then divided into a Foreground and a Background Branch based on their corresponding saliency scores $S$. The details of SDM are explained in Section \ref{sec:SDM}. Subsequently, we propose a three-stage Hierarchical Refinement Module (HRM) to further enhance the fine-grained localization and semantic precision of disentangled correlation maps $C_f$ and $C_b$ separately via Pixel-wise, Category-wise, and Semantic-wise Refinement. Detailed explanations of HRM are in Section \ref{sec:HRM}. Afterwards, the refined features ($C''_f$ and $C''_b$) from the foreground and background branches are integrated through a weighted feature aggregation block to produce aggregated correlation maps $\widetilde{C} \in \mathbb{R}^{H \times W \times N_C \times D}$. Finally, it produces the final mask predictions $\boldsymbol{\hat{y}}$ through an upsampling decoder. \par

\vspace{-0.5\baselineskip}
\subsection{Saliency-aware Disentanglement Module} \label{sec:SDM}
We design the SDM to address the inherent \textbf{\textit{Foreground Bias}} in VLMs. It uses GradCAM \citep{selvaraju2017grad} to generate per-class saliency maps from cross-attention. Traditional saliency-based methods merely focus on improving model efficiency through token pruning. However, we incorporate saliency as an additional visual cue to perform complementary feature learning. The saliency is consistent with prior work that reflects semantic contribution but with a slight modification: we obtain saliency maps for each class instead of a single saliency map for all classes. In our method, saliency therefore represents regions of semantic details and informative structures, not merely regions of model confidence. Saliency provides a scalar importance score for disentanglement. It captures where meaningful evidence appears, while correlation encodes the semantic cues present in that region. The gradient-based saliency generation is essential for robust disentanglement. \par

\begin{figure*}[t]
\centering
\begin{overpic}[width=0.65\linewidth]{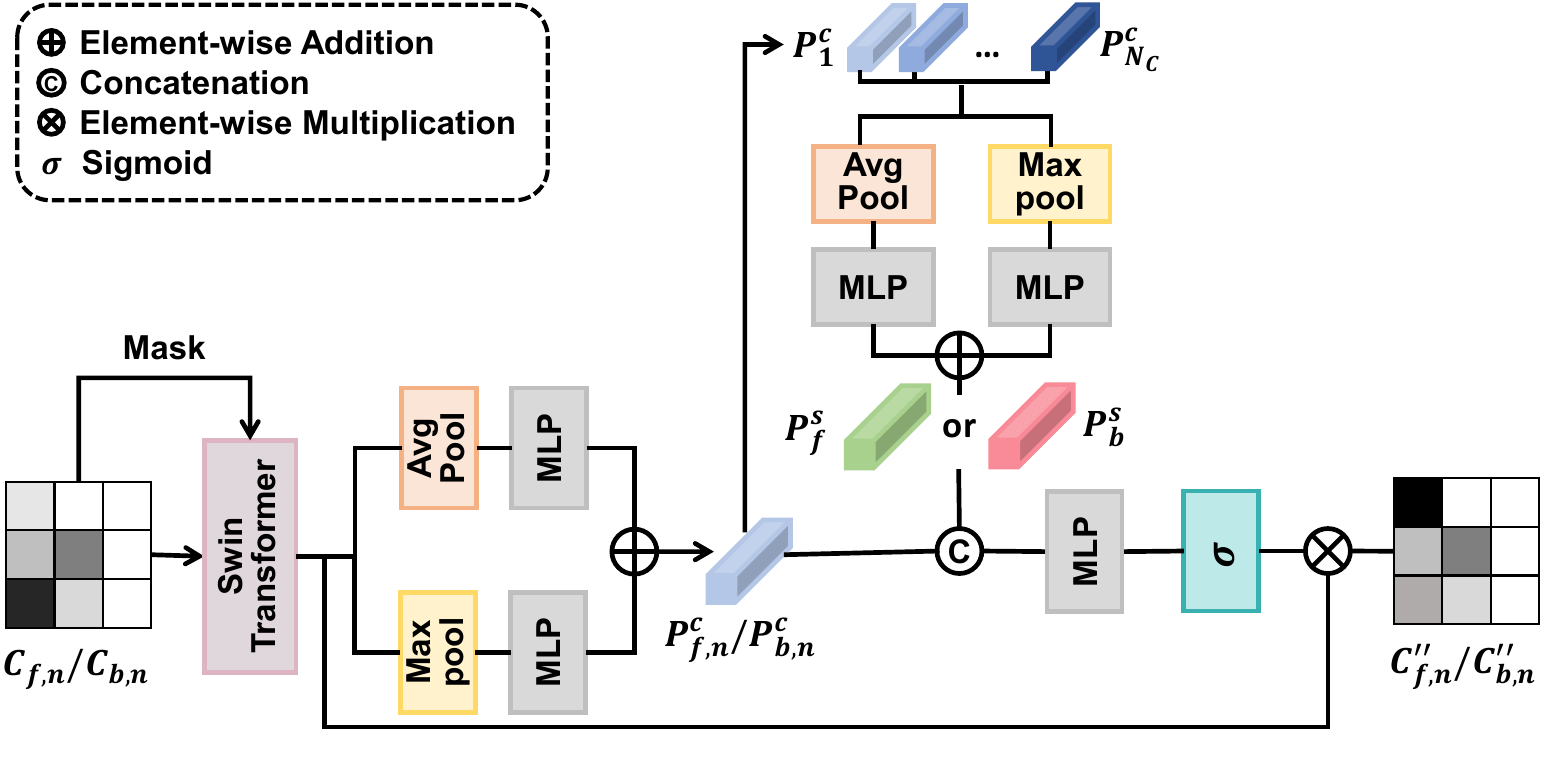}
  \put(6, -4){\small (a) \textbf{Pixel-wise Refinement}}       % left column caption
  \put(46, -4){\small (b) \textbf{Category-} and \textbf{Semantic-wise Refinement}}  % middle column
\end{overpic}
\vspace{4mm}
\caption{\textbf{Illustration of Hierarchical Refinement Module (HRM).}}
\vspace{-1.2\baselineskip}
\label{fig:HRM}
\end{figure*}

\noindent \textbf{Saliency Map Generation.} Given image $F_v$ and text embeddings $F_t$, we employ cross-attention layers where text embeddings serve as the query and image embeddings serve as the key/value. The intermediate attention map $A$ captures image-text correspondences; however, the attentions typically exhibit scattered and spatially diffuse activations due to their overly broad receptive fields, which are inherent in global modeling \citep{wang2025iterprime}. To address this limitation, we selectively suppress less-relevant regions within the attention map $A$ and enhance its spatial precision via gradient-based reweighting. This objective is achieved through an auxiliary Image-Text Matching (ITM) loss $\mathcal{L}_\text{itm}$ \citep{li2021align} that provides explicit supervision for localization. Specifically, we append an auxiliary regression head to classify whether each image-text pair is matched. The ITM loss is formulated as:
\begin{equation} \label{eq:ITM}
\mathcal{L}_\text{itm} = \mathbb{E}_{(v, t)\sim D} \mathcal{H}(\boldsymbol{\hat{y}}^{itm}_{(v, t)}, \boldsymbol{y}^{itm})
\end{equation}
where $\mathcal{H}$ is the cross-entropy loss, and ground-truth labels $\boldsymbol{y}^{itm}$ are one-hot vectors obtained from segmentation masks. During inference, we use the regressed image-text matching scores $\boldsymbol{\hat{y}}^{itm}_{(v, t)}$ to generate gradients, so no ground truth or class labels are required, thereby avoiding data leakage. Afterwards, we compute the gradient of $\mathcal{L}_\text{itm}$ and let the attention maps narrowly focus on the most discriminative regions through GradCAM-style re-weighting to produce the saliency map $S_n$ for the n-th class, following PnP-OVSS:
\begin{equation} \label{eq:saliency}
S_n = \max\left(0,\; \frac{\partial \mathcal{L}_\text{itm}}{\partial A_n}\right) \otimes A_n
\end{equation}
where $\otimes$ represents element-wise multiplication. Eq. \ref{eq:saliency} presents the general formulation, while during inference the loss term is replaced by the regressed matching scores and more details are in the appendix. Note that the goal of this auxiliary sharpening loss is to enhance contrast and to back-propagate gradients for saliency generation. The quality of disentanglement is not heavily dependent on the quality and accuracy of it. Instead, we rely on the image-text attention maps $A$, which provide robust localization cues and are continuously optimized during training through segmentation objectives, thereby stabilizing and enhancing their even if the auxiliary ITM supervision is imperfect. \par

\noindent \textbf{Foreground/Background Token Selection.} Unlike traditional approaches that treat all visual tokens uniformly, we propose a dual-branch mechanism to disentangle all visual tokens into the Foreground and Background Branches by saliency-aware feature decomposition. It enables the model to explicitly distinguish between foreground and background regions, focusing on the distinctive characteristics of both features. This architectural decoupling directly mitigates \textbf{\textit{Foreground Bias}} in VLMs caused by their object-centric focus during pre-training. It enables each branch to develop domain-specific representations. \par

Specifically, for the n-th class, we select the top-k visual tokens in the correlation maps $C_n$ corresponding to their saliency scores $S_n$ through a binary mask as foreground correlation maps $C_{f,n}$ while the remaining are designated as background correlation maps $C_{b,n}$. These disentangled maps are then processed through two specialized branches: a Foreground Branch that models salient foreground features, and a Background Branch that captures contextual background information. \par

\vspace{-0.5\baselineskip}
\subsection{Hierarchical Refinement Module} \label{sec:HRM}
Existing SOTA methods often struggle with effectively capturing the boundary details in complex scenes. To address this challenge, we perform hierarchical refinement to update the correlation maps $C$ across three distinct levels: (1) \textbf{Pixel-wise Refinement}, which focuses on achieving precise spatial localization; (2) \textbf{Category-wise Refinement}, aimed at enhancing channel-wise coherence for each class; and (3) \textbf{Semantic-wise Refinement}, which captures semantic consistency within broader foreground/background groupings. The hierarchical design preserves fine-grained spatial details while capturing cross-channel context, leading to improved segmentation precision in complex visual environments. \par

\noindent \textbf{Pixel-wise Refinement.} As shown in Fig. \ref{fig:HRM}, Pixel-wise Refinement takes foreground correlation maps $C_{f}$ and background correlation maps $C_{b}$ as input. The Pixel-wise Refinement block is applied for spatial aggregation based on the Swin Transformer \citep{liu2021swin} block with key modifications. Instead of performing cross-attention across all tokens, we apply masked attention within the block in two branches to focus solely on foreground and background embeddings, respectively. It consists of 2 blocks: the first block implements attention within a local window, while the second block employs shifted window-based attention to enhance global context integration. The outputs are well-refined correlation maps ($C'_{f}$ and $C'_{b}$) after pixel-wise spatial aggregation, suppressing noise in image-text correlations. \par

\noindent \textbf{Category-wise Refinement.} Subsequent to Pixel-wise Refinement, Category-wise Refinement is applied to consider category-specific cross-covariance across feature channels. Given the pixel-refined correlation maps $C'_{f}$ and $C'_{b}$, we apply 2D global average pooling and max pooling in parallel to extract both the spatial extent of target objects and discriminative clues across channel dimensions. The resulting pooled features are independently processed by 2 MLPs, and their outputs are combined through element-wise addition to generate the category prototype $P^c \in \mathbb{R}^{1 \times N_C \times D}$. Note that the $P^c$ is class-specific, and $P^c_{f, n}$ and $P^c_{b, n}$ represent the category prototype for the Foreground and Background Branches of the n-th class. \par

\noindent \textbf{Semantic-wise Refinement.} While Category-wise Refinement enhances channel-wise coherence, it is class-specific and may overlook broader contextual cues, e.g., surrounding environments or overall scenes among all classes. To improve semantic understanding of the relationship among all salient objects for foreground and all semantic regions in the environments for background, we design the Semantic-wise Refinement. This additional refinement block considers coarse-grained scene context across all classes, leading to more robust and generalized representations. \par

\begin{table*}[t]
\centering
\resizebox{\textwidth}{!}{
\begin{tabular}{lccccccccccc}
\toprule
Model & VLM & Additional Backbone & Training Dataset & Additional Dataset & A-847 & PC-459 & A-150 & PC-59 & PAS-20 & PAS-20$^{b}$ \\
\midrule
LSeg \textcolor{gray}{[arXiv21]} & CLIP ViT-B/32 & ResNet-101 & PASCAL VOC-15 & $\newcrossmark$ & - & - & - & - & 47.4 & - \\
LSeg+ \textcolor{gray}{[ECCV22]} & ALIGN & ResNet-101 & COCO-Stuff & $\newcrossmark$ & 2.5 & 5.2 & 13.0 & 36 & - & 59.0 \\
ZegFormer \textcolor{gray}{[CVPR22]} & CLIP ViT-B/16 & ResNet-101 & COCO-Stuff-156 & $\newcrossmark$ & 4.9 & 9.1 & 16.9 & 42.8 & 86.2 & 62.7 \\
ZegFormer \textcolor{gray}{[CVPR22]} & CLIP ViT-B/16 & ResNet-101 & COCO-Stuff & $\newcrossmark$ & 5.6 & 10.4 & 18.0 & 45.5 & 89.5 & 65.5 \\
ZSseg \textcolor{gray}{[ECCV22]} & CLIP ViT-B/16 & ResNet-101 & COCO-Stuff & $\newcrossmark$ & 7.0 & - & 20.5 & 47.7 & 88.4 & - \\
OpenSeg \textcolor{gray}{[ECCV22]} & ALIGN & ResNet-101 & COCO Panoptic & $\checkmark$ & 4.4 & 7.9 & 17.5 & 40.1 & - & 63.8 \\
OVSeg \textcolor{gray}{[CVPR23]} & CLIP ViT-B/16 & ResNet-101 & COCO-Stuff & $\checkmark$ & 7.1 & 11.0 & 24.8 & 53.3 & 92.6 & - \\
ZegCLIP \textcolor{gray}{[CVPR23]} & CLIP ViT-B/16 & - & COCO-Stuff-156 & $\newcrossmark$ & - & - & - & 41.2 & 93.6 & - \\
SAN \textcolor{gray}{[CVPR23]} & CLIP ViT-B/16 & - & COCO-Stuff & $\newcrossmark$ & 10.1 & 12.6 & 27.5 & 53.8 & 94.0 & - \\
EBSeg \textcolor{gray}{[CVPR24]} & CLIP ViT-B/16 & - & COCO-Stuff & $\newcrossmark$ & 11.1 & 17.3 & 30.0 & 56.7 & 94.6 & - \\
SED \textcolor{gray}{[CVPR24]} & ConvNeXt-B & - & COCO-Stuff & $\newcrossmark$ & 11.4 & 18.6 & 31.6 & 57.3 & 94.4 & - \\
CAT-Seg \textcolor{gray}{[CVPR24]} & CLIP ViT-B/16 & - & COCO-Stuff & $\newcrossmark$ & \secondbest{12.0} & 19.0 & 31.8 & 57.5 & 94.6 & \secondbest{77.3} \\
DPSeg \textcolor{gray}{[CVPR25]} & CLIP ViT-B/16 & - & COCO-Stuff & $\newcrossmark$ & \secondbest{12.0} & \secondbest{19.5} & \secondbest{32.9} & \secondbest{58.1} & \secondbest{96.0} & - \\
\rowcolor{gray!10}
\textbf{DiSa} & CLIP ViT-B/16 & - & COCO-Stuff & $\newcrossmark$ & \best{12.6} & \best{20.3} & \best{33.7} & \best{59.3} & \best{97.0} & \best{79.9} \\
& & & & & \best{(+0.6)} & \best{(+0.8)} & \best{(+0.8)} & \best{(+1.2)} & \best{(+1.0)} & \best{(+2.6)} \\
\midrule
LSeg \textcolor{gray}{[arXiv21]} & CLIP ViT-B/32 & ViT-L/16 & PASCAL VOC-15 & $\newcrossmark$ & - & - & - & - & 52.3 & - \\
OpenSeg \textcolor{gray}{[ECCV22]} & ALIGN & Eff-B7 & COCO Panoptic & $\checkmark$ & 8.1 & 11.5 & 26.4 & 44.8 & - & 70.2 \\
OVSeg \textcolor{gray}{[CVPR23]} & CLIP ViT-L/14 & Swin-B & COCO-Stuff & $\checkmark$ & 9.0 & 12.4 & 29.6 & 55.7 & 94.5 & - \\
SAN \textcolor{gray}{[CVPR23]} & CLIP ViT-L/14 & - & COCO-Stuff & $\newcrossmark$ & 12.4 & 15.7 & 32.1 & 57.7 & 94.6 & - \\
ODISE \textcolor{gray}{[CVPR23]} & CLIP ViT-L/14 & Stable Diffusion & COCO-Stuff & $\newcrossmark$ & 11.1 & 14.5 & 29.9 & 57.3 & - & - \\
SCAN \textcolor{gray}{[CVPR24]} & CLIP ViT-L/14 & - & COCO-Stuff & $\newcrossmark$ & 14.0 & 16.7 & 33.5 & 59.3 & 97.2 & - \\
EBSeg \textcolor{gray}{[CVPR24]} & CLIP ViT-L/14 & - & COCO-Stuff & $\newcrossmark$ & 13.7 & 21.0 & 32.8 & 60.2 & 96.4 & - \\
SED \textcolor{gray}{[CVPR24]} & ConvNeXt-L & - & COCO-Stuff & $\newcrossmark$ & 13.9 & 22.6 & 35.2 & 60.6 & 96.1 & - \\
CAT-Seg \textcolor{gray}{[CVPR24]} & CLIP ViT-L/14 & - & COCO-Stuff & $\newcrossmark$ & \secondbest{16.0} & \secondbest{23.8} & \secondbest{37.9} & \secondbest{63.3} & 97.0 & \secondbest{82.5} \\
DPSeg \textcolor{gray}{[CVPR25]} & CLIP ViT-L/14 & - & COCO-Stuff & $\newcrossmark$ & 14.9 & 23.5 & 36.4 & 62.0 & \secondbest{97.4} & - \\
\rowcolor{gray!10}
\textbf{DiSa} & CLIP ViT-L/14 & - & COCO-Stuff & $\newcrossmark$ & \best{16.3} & \best{24.9} & \best{38.9} & \best{64.7} & \best{98.7} & \best{84.7} \\
 & & & & & \best{(+0.3)} & \best{(+1.1)} & \best{(+1.0)} & \best{(+1.4)} & \best{(+1.3)} & \best{(+2.2)} \\
\bottomrule
\end{tabular}}
\caption{\textbf{Quantitative results on 6 benchmarks.} The best-performing results are presented in \textbf{bold}, while the second-best results are \underline{underlined}. Improvements over the second-best are in \textbf{bold}.}
\vspace{-1.4\baselineskip}
\label{tab:main_results}
\end{table*}

Similar to Category-wise Refinement, we apply 1D global average pooling and max pooling layers among category prototypes of all classes $\{P^c_i, i=1, 2,..., N_c\}$ to extract semantic prototypes $P^s \in \mathbb{R}^{1 \times 1 \times D}$. Note that $P^{s}_f$ and $P^s_b$ are class-agnostic and shared among all classes within the Foreground and Background Branches. After extracting both category $P^c$ and semantic prototypes $P^s$, we update the pixel-refined correlation maps $C'_{f}$ and $C'_{b}$ by aggregating these channel-wise cues. Specifically, for the n-th class, we fuse both $P^c_n$ and $P^s$ by concatenation. Note that the semantic prototype used is selected based on the branch: $P^s_f$ for the Foreground Branch and $P^s_b$ for the Background Branch. This fused output is then element-wise multiplied by pixel-refined correlation embeddings $C'$, followed by a sigmoid activation:
\begin{equation} \label{eq:hrm_fore}
C''_{f,i} = C'_{f,i} \otimes \sigma(\text{MLP}(\text{Concat}(P^c_{f,i},  P^s_f)))
\end{equation}
\begin{equation} \label{eq:hrm_back}
C''_{b,i} = C'_{b,i} \otimes \sigma(\text{MLP}(\text{Concat}(P^c_{b,i},  P^s_b)))
\end{equation}
where $i=1,2,...,N_C$, $\text{Concat}$ is concatenation, $\sigma(\cdot)$ is sigmoid, and $\otimes$ refers to element-wise multiplication. \par

By hierarchically refining correlation maps across pixel-, category-, and semantic-levels, HRM makes the saliency-aware correlation representations more informative and enhances the accuracy, leading to fine-grained spatial precision in downstream segmentation tasks. \par

\subsection{Foreground and Background Aggregation}
To uniformly model all pixels after capturing foreground- and background-specific features, we use learnable weights by gating mechanism to aggregate disentangled refined correlation maps ($C''_{f,n}$ and $C''_{b,n}$) for the n-th class with previous binary mask (reorganization based on mask indices). We further employ a Swin Transformer block to mitigate potential misalignment between dual branches, and the aggregated correlation maps are denoted as $\widetilde{C}$. Finally, $\widetilde{C}$ serves as visual guidance and are fed into an upsampling decoder, along with image embeddings $F_v$ from the CLIP image encoder, to generate the final mask predictions $\boldsymbol{\hat{y}}$. \par
\vspace{-0.5\baselineskip}
\section{Experiments}  \label{sec:expts}
\vspace{-0.5\baselineskip}
\subsection{Datasets}
\vspace{-0.5\baselineskip}
Our experiments are trained on COCO-Stuff \citep{caesar2018coco} and evaluated on 6 large-scale semantic segmentation datasets %\footnote{Datasets were downloaded and evaluated solely by Lehigh University.}
. \textbf{ADE20K} \citep{zhou2019semantic} is a large-scale benchmark for semantic segmentation with 2000 validation images, supporting two evaluation protocols: ADE-150 with 150 categories, and ADE-847 with extended 847 classes. \textbf{PASCAL-VOC} \citep{everingham2010pascal} is a widely used dataset containing 1,500 validation images with 20 foreground categories, referred as PAS-20. Another evaluation protocol PAS-20$^{b}$ \citep{ghiasi2022scaling} with one extra class for background is also included. \textbf{PASCAL-Context} \citep{mottaghi2014role} extends PASCAL VOC, supporting 2 evaluation protocols: PC-59 with 59 labeled classes and PC-459 with 459 categories. \par

\begin{figure*}[t]
\centering
\includegraphics[width=0.85\linewidth]{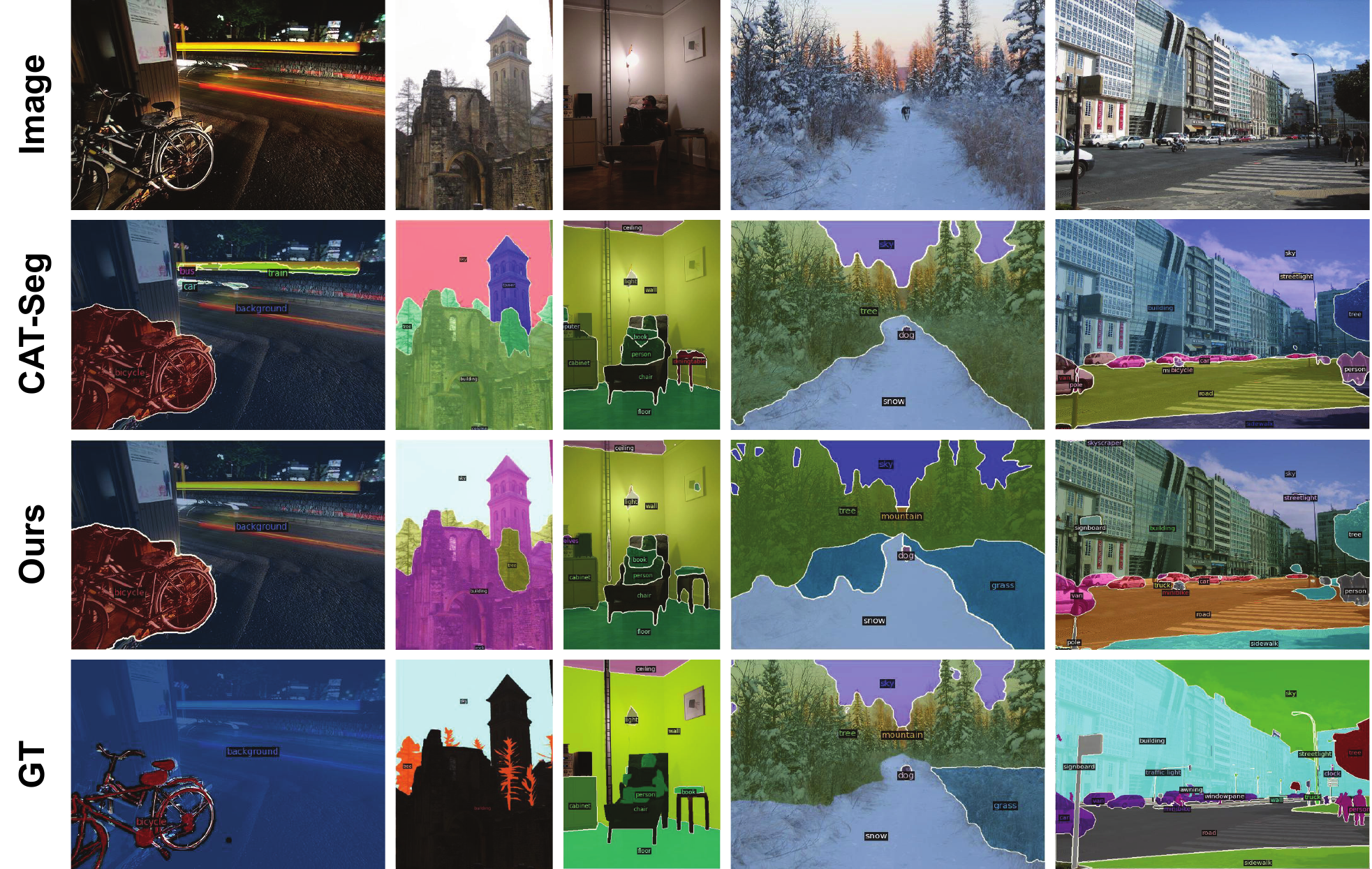}
\caption{\textbf{Qualitative results compared to CAT-Seg.} DiSa produces
more accurate predictions of small objects and visually-similar regions compared to existing SOTA methods. More qualitative results are in the appendix.}
\vspace{-0.8\baselineskip}
\label{fig:vis}
\end{figure*}

\vspace{-0.5\baselineskip}
\subsection{Implementation Details}
\vspace{-0.5\baselineskip}
We implement our work using PyTorch \citep{paszke2019pytorch} and Detectron2 \citep{wu2019detectron2}. The loss function is a weighted sum of cross-entropy loss (1) and the ITM loss (0.2). We set $D=128$, and training resolution to $384\times384$. The $k$ value for selecting foreground tokens is 96. The decoder consists of 2 transposed convolution layers that take $\widetilde{C}$ and $F_v$ as inputs. Following CAT-Seg \citep{cho2024cat}, we fine-tune query/key in attention of CLIP image and text encoders. We train the model using the AdamW optimizer \citep{loshchilov2017decoupled} with batch size 2. The learning rate is 2e-4 for our designed modules and 2e-6 for CLIP encoders. We use 2 NVIDIA RTX A5000 GPUs for training. All models are trained for 80,000 iterations. \par

\vspace{-0.5\baselineskip}
\subsection{Quantitative Results}
\vspace{-0.5\baselineskip}
Table \ref{tab:main_results} demonstrates quantitative results on standard OVSS datasets \citep{zhou2019semantic,everingham2010pascal,mottaghi2014role}. We compare existing works, LSeg \citep{li2022language}, LSeg+ \citep{ghiasi2022scaling}, ZegFormer \citep{ding2022decoupling}, ZSseg \citep{xu2022simple}, OpenSeg \citep{ghiasi2022scaling}, OVSeg \citep{liang2023open}, ZegCLIP \citep{zhou2023zegclip}, SAN \citep{xu2023side}, ODISE \citep{xu2023open}, SCAN \citep{liu2024open}, EBSeg \citep{shan2024open}, SED \citep{xie2024sed}, CAT-Seg \citep{cho2024cat}, and DPSeg \citep{zhao2025dpseg} with similar-scale VLMs. Note that we adopt the DPSeg \citep{zhao2025dpseg} inference I model for fair comparison. Unlike some prior works, our model does not leverage any additional datasets or backbones. \par

\begin{figure}[t]
\centering
\vspace{0.8\baselineskip}
\begin{overpic}[width=0.92\linewidth]{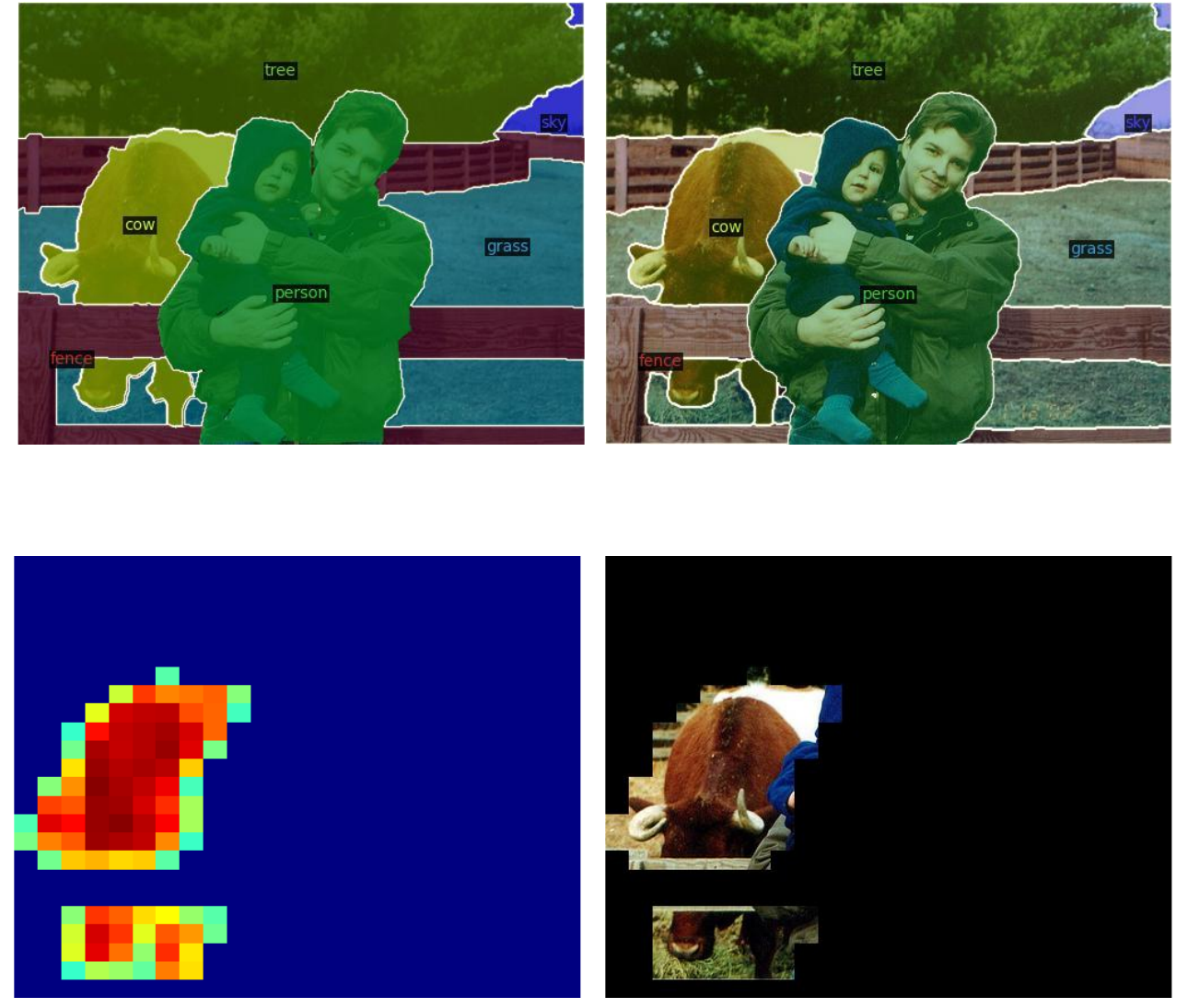}
  \put(10, 42){\small (a) Ground Truth}       % left column caption
  \put(62, 42){\small (b) Prediction}  % middle column
  \put(5, -4){\small (c) Foreground saliency}       % left column caption
  \put(55, -4){\small (d) Foreground tokens}  % middle column
\end{overpic}
\vspace{0.8\baselineskip}
\caption{\textbf{Qualitative comparison of saliency and predictions.}}
\label{fig:feature}
\vspace{-1.8\baselineskip}
\end{figure}
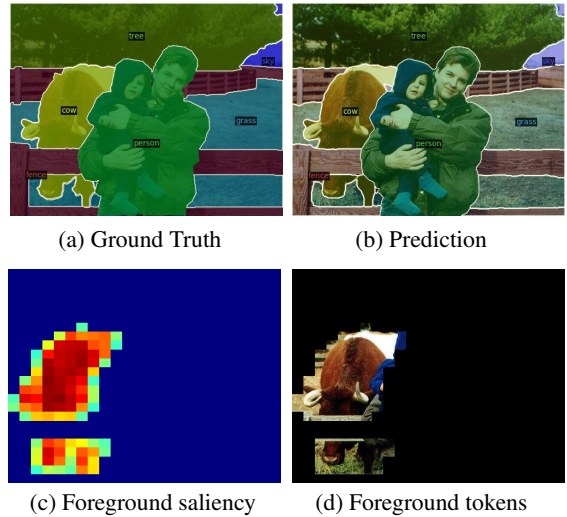

Our method, DiSa, achieves consistent and significant gains across all benchmarks, in both base-VLM and large-VLM settings. As shown in Table \ref{tab:main_results}, in the base-VLM configuration, DiSa outperforms prior SOTA approaches with the improvements of +0.6\%, +0.8\%, +0.8\%, +1.2\%, +1.0\%, and +2.6\% mIoU (with an average performance gain of +1.2\% mIoU), on A-847, PC-459, A-150, PC-59, PAS-20, and PAS-20$^{b}$, respectively. For the large-VLM configuration, DiSa outperforms the best prior SOTA by +0.3\%, +1.1\%, +1.0\%, +1.4\%, +1.3\%, and +2.2\% mIoU with an average gain of 1.2\% mIoU among all datasets. Note that DiSa has the most significant relative performance gains on PAS-20$^{b}$, demonstrating that it mitigates \textbf{\textit{Foreground Bias}}. These gains are not only statistically meaningful but also practically significant given the performance saturation observed in this task. \par

We attribute the leading performance of DiSa to two factors: (1) Our Saliency-aware Disentanglement enhances context-aware features while preserving semantic coherence. It mitigates \textbf{\textit{Foreground Bias}}, as demonstrated by significant improvements on PAS-20$^{b}$, which includes background classes. (2) Hierarchical Refinement Module yields accurate and robust boundaries via multi-level refinement, contributing to consistent performance gains across all datasets. \par

\begin{table}[t]
\centering
\resizebox{0.92\linewidth}{!}{%
\begin{tabular}{lcccc}
\toprule
\textbf{Model} & \textbf{\# Params. (M)} & \textbf{GFLOPs} & \textbf{Inference time (s)} \\
\midrule
ZegFormer & 531.2 & 19{,}425.6 & 3.10 \\
ZSseg & 530.8 & 22{,}302.1 & 3.11 \\
OVSeg & 532.6 & 19{,}345.6 & 2.98 \\
CAT-Seg & 433.7 & 2{,}121.1 & 0.78 \\
% \rowcolor{gray!10}
ESC-Net & 451.3 & 2{,}203.5 & 0.76 \\
\midrule
Ours & 456.2 & 2{,}287.3 & 0.69 \\
\bottomrule
\end{tabular}
}
\caption{\textbf{Model complexity comparison.} We use CLIP ViT-B/16 for VLM and one single A6000 GPU for fair comparison.}
\vspace{-1.2\baselineskip}
\label{tab:efficiency}
\end{table}

\vspace{-0.6\baselineskip}
\subsection{Qualitative Results}
\vspace{-0.6\baselineskip}
We evaluate qualitative results of our method with CAT-Seg \citep{cho2024cat} using default settings in Fig. \ref{fig:vis}. We present diverse scenarios, including crowded background (columns 1\&5) and visually similar classes (rows 2-4). CAT-Seg struggles to handle complex foreground-background relations and locate accurate boundaries. For example, in column 1, the background is misclassified as “train”. Similarly, in columns 2\&4, CAT-Seg produces ambiguous boundaries between visually similar categories (e.g., “snow” and “grass”), reflecting its limited spatial localization. In contrast, DiSa preserves object integrity in crowded scenes, demonstrating superior robustness in challenging scenarios. \par

We present visualizations of foreground saliency and image tokens of a specific class “cow” in Fig. \ref{fig:feature}. We observe that foreground and background tokens of the partially occluded cow are identified and not suppressed by other categories. It is consistent with our design, yielding sharper boundaries. \par

\begin{table}[t]
\centering
\resizebox{\linewidth}{!}{%
\begin{tabular}{lcccccc}
\toprule
\textbf{Decomposition} & \textbf{A-847} & \textbf{PC-459} & \textbf{A-150} & \textbf{PC-59} & \textbf{PAS-20} & \textbf{PAS-20\textsuperscript{b}} \\
\midrule
(I) No disentanglement & 11.4 & 18.8 & 30.9 & 56.3 & 94.5 & 76.1 \\
(II) Token-level & \secondbest{12.1} & \secondbest{19.9} & \secondbest{32.6} & \secondbest{58.2} & 94.5 & \secondbest{78.9} \\
(III) Class-level & 11.5 & 19.5 & 31.1 & \secondbest{58.2} & \secondbest{95.1} & 78.5 \\
\midrule
(IV) DiSa w/o HRM & 11.7 & 18.7 & 31.2 & 57.7 & 95.3 & 78.5 \\
(V) (IV) + Pixel & 12.4 & 19.5 & 32.3 & 58.8 & 95.9 & 78.9 \\
(VI) (IV) + Category & 11.8 & 19.1 & 32.0 & 57.9 & 95.4 & 78.3 \\
(VII) (V) + Category & \secondbest{12.5} & \secondbest{19.8} & \secondbest{33.1} & \secondbest{59.0} & \secondbest{96.6} & \secondbest{79.1} \\
\midrule
(VIII) \textbf{Ours} & \best{12.6} & \best{20.3} & \best{33.7} & \best{59.3} & \best{97.0} & \best{79.9} \\
\bottomrule
\end{tabular}
}
\caption{\textbf{Ablation study for various design choices.} CLIP ViT-B/16 is used as VLM for ablation.}
\vspace{-1.8\baselineskip}
\label{table:ablation}
\end{table}

\vspace{-0.5\baselineskip}
\subsection{Model Efficiency Analysis}\label{sec:complexity}
\vspace{-0.5\baselineskip}
We further conduct the model efficiency analysis (parameter size and GFLOPS) on all 6 datasets in Table \ref{tab:efficiency}. We report the mean run-time on five datasets, following \cite{cho2024cat}. Notably, ZegFormer \citep{ding2022decoupling}, ZSseg \citep{xu2022simple}, and OVSeg \citep{liang2023open} rely on large-scale backbones and complex vision-language fusion modules, with more than 530M parameters and 19k GFLOPs. In contrast, our model significantly reduces inference cost to 2k GFLOPs while maintaining a competitive parameter count of 456M. Although slightly larger than CAT-Seg \citep{cho2024cat} and ESC-Net \citep{lee2025effective}, our framework achieves comparable efficiency and remains lightweight compared to other CLIP-based methods. Our gradient-based computation is lightweight because the gradient is only back-propagated to attention maps, not through the entire network. This is a very short computational path with minimal overhead of the saliency process. \par

\vspace{-0.5\baselineskip}
\subsection{Ablation Study}
\vspace{-0.5\baselineskip}
\noindent \textbf{Design choices for disentanglement.} To validate the impact of our saliency-aware disentanglement, we compare our \textbf{(I)} baseline (DiSa without FG/BG disentanglement which only contains single branch for all tokens) with 2 other designs in Table \ref{table:ablation}: \textbf{(II)} token-level, following PraNet \citep{hu2025pranet} to decouple foreground/background features via FB/BG supervision, \textbf{(III)} class-level, leveraging LLMs \citep{achiam2023gpt} for a pre-defined taxonomy and separating all classes into 2 branches. As shown in the table, token-level disentanglement \textbf{(II)} achieves marginal improvements over the baseline (with an average gain of 0.66\%). Class-level disentanglement \textbf{(III)} slightly improves on some benchmarks, likely due to its rigidity in adapting to varying scene contexts. In contrast, our proposed saliency-aware disentanglement \textbf{(VIII)} consistently outperforms other decompositions by 1.1\% (token-level) and 1.5\% (class-level) on average. Notably, it yields a substantial improvement (2.6\%) on PAS-20$^{b}$, effectively alleviating \textbf{\textit{Foreground Bias}}. \par

\noindent \textbf{Component analysis for HRM.} To validate the effectiveness of HRM, we further evaluate the performance gain of four variants \textbf{(IV-VII)} by gradually adding the components in Table \ref{table:ablation}. Specifically, they are: \textbf{(IV)} baseline (DiSa without HRM), \textbf{(V)} adding Pixel-wise Refinement to \textbf{(IV)}, \textbf{(VI)} adding Category-wise Refinement to \textbf{(IV)}, \textbf{(VII)} adding Category-wise Refinement to \textbf{(V)}, and \textbf{(VIII)} employing all designed components. Introducing Pixel-wise Refinement \textbf{(V)} improves the average mIoU by 0.78\%. Adding Category-wise Refinement \textbf{(VII)} further boosts performance by capturing channel-wise category semantics, with an average gain of 1\% over the baseline. Finally, incorporating Semantic-wise Refinement \textbf{(VIII)} yields the highest overall performance (1.62\% on average). It demonstrates that HRM and multi-level refinement are essential for mitigating \textbf{\textit{Limited Spatial Localization}} and semantic coherence.  \par

\vspace{-0.8\baselineskip}
\section{Conclusion}  \label{sec:conclusion}
\vspace{-0.3\baselineskip}
In this paper, we propose DiSa, a novel Saliency-aware Foreground-background Disentangled framework for open-vocabulary semantic segmentation. To address the \textbf{\textit{Foreground Bias}} and \textbf{\textit{Limited Spatial Localization}} limitations inherent in VLMs, we propose a Saliency-aware Disentanglement Module (SDM), which performs adaptive foreground-background decomposition based on saliency cues, enabling context-dependent ensemble feature learning. Additionally, by integrating a Hierarchical Refinement Module (HRM), DiSa yields fine-grained spatial localization through Pixel-, Category-, and Semantic-wise Refinement. Extensive experimental results on six large-scale datasets demonstrate the effectiveness of our model. Our observations and novel design shift the paradigm and suggest a promising direction for future research. \par

\section*{Impact Statement}
Our work aims to advance open-vocabulary semantic segmentation by explicitly disentangling foreground and background semantics when leveraging vision-language priors. Concretely, we separate class-related evidence into foreground-focused cues and context/background cues, and then reason over each stream (and their interaction) to better capture both object identity and scene context. This foreground/background disentanglement helps alleviate the common bias of pretrained VLMs toward salient object regions while under-representing contextual semantics, improving generalization to novel concepts and reducing dependence on dense pixel-level annotation. \par

\bibliography{main}

@String(AAAI = {AAAI})

@inproceedings{caesar2018coco,
  title={Coco-stuff: Thing and stuff classes in context},
  author={Caesar, Holger and Uijlings, Jasper and Ferrari, Vittorio},
  booktitle={Proceedings of the IEEE conference on computer vision and pattern recognition},
  pages={1209--1218},
  year={2018}
}

@article{zhou2019semantic,
  title={Semantic understanding of scenes through the ade20k dataset},
  author={Zhou, Bolei and Zhao, Hang and Puig, Xavier and Xiao, Tete and Fidler, Sanja and Barriuso, Adela and Torralba, Antonio},
  journal={International Journal of Computer Vision},
  volume={127},
  pages={302--321},
  year={2019},
  publisher={Springer}
}

@article{everingham2010pascal,
  title={The pascal visual object classes (voc) challenge},
  author={Everingham, Mark and Van Gool, Luc and Williams, Christopher KI and Winn, John and Zisserman, Andrew},
  journal={International journal of computer vision},
  volume={88},
  pages={303--338},
  year={2010},
  publisher={Springer}
}

@inproceedings{mottaghi2014role,
  title={The role of context for object detection and semantic segmentation in the wild},
  author={Mottaghi, Roozbeh and Chen, Xianjie and Liu, Xiaobai and Cho, Nam-Gyu and Lee, Seong-Whan and Fidler, Sanja and Urtasun, Raquel and Yuille, Alan},
  booktitle={Proceedings of the IEEE conference on computer vision and pattern recognition},
  pages={891--898},
  year={2014}
}

@article{li2024densevlm,
  title={DenseVLM: A Retrieval and Decoupled Alignment Framework for Open-Vocabulary Dense Prediction},
  author={Li, Yunheng and Li, Yuxuan and Zeng, Quansheng and Wang, Wenhai and Hou, Qibin and Cheng, Ming-Ming},
  journal={arXiv preprint arXiv:2412.06244},
  year={2024}
}

@inproceedings{zhang2023simple,
  title={A simple framework for open-vocabulary segmentation and detection},
  author={Zhang, Hao and Li, Feng and Zou, Xueyan and Liu, Shilong and Li, Chunyuan and Yang, Jianwei and Zhang, Lei},
  booktitle={Proceedings of the IEEE/CVF International Conference on Computer Vision},
  pages={1020--1031},
  year={2023}
}

@inproceedings{zhong2022regionclip,
  title={Regionclip: Region-based language-image pretraining},
  author={Zhong, Yiwu and Yang, Jianwei and Zhang, Pengchuan and Li, Chunyuan and Codella, Noel and Li, Liunian Harold and Zhou, Luowei and Dai, Xiyang and Yuan, Lu and Li, Yin and others},
  booktitle={Proceedings of the IEEE/CVF conference on computer vision and pattern recognition},
  pages={16793--16803},
  year={2022}
}

@inproceedings{zhou2022extract,
  title={Extract free dense labels from clip},
  author={Zhou, Chong and Loy, Chen Change and Dai, Bo},
  booktitle={European Conference on Computer Vision},
  pages={696--712},
  year={2022},
  organization={Springer}
}

@inproceedings{cho2024cat,
  title={Cat-seg: Cost aggregation for open-vocabulary semantic segmentation},
  author={Cho, Seokju and Shin, Heeseong and Hong, Sunghwan and Arnab, Anurag and Seo, Paul Hongsuck and Kim, Seungryong},
  booktitle={Proceedings of the IEEE/CVF Conference on Computer Vision and Pattern Recognition},
  pages={4113--4123},
  year={2024}
}

@inproceedings{xian2019semantic,
  title={Semantic projection network for zero-and few-label semantic segmentation},
  author={Xian, Yongqin and Choudhury, Subhabrata and He, Yang and Schiele, Bernt and Akata, Zeynep},
  booktitle={Proceedings of the IEEE/CVF Conference on Computer Vision and Pattern Recognition},
  pages={8256--8265},
  year={2019}
}

@article{bucher2019zero,
  title={Zero-shot semantic segmentation},
  author={Bucher, Maxime and Vu, Tuan-Hung and Cord, Matthieu and P{\'e}rez, Patrick},
  journal={Advances in Neural Information Processing Systems},
  volume={32},
  year={2019}
}

@article{li2022language,
  title={Language-driven semantic segmentation},
  author={Li, Boyi and Weinberger, Kilian Q and Belongie, Serge and Koltun, Vladlen and Ranftl, Ren{\'e}},
  journal={arXiv preprint arXiv:2201.03546},
  year={2022}
}

@inproceedings{ghiasi2022scaling,
  title={Scaling open-vocabulary image segmentation with image-level labels},
  author={Ghiasi, Golnaz and Gu, Xiuye and Cui, Yin and Lin, Tsung-Yi},
  booktitle={European conference on computer vision},
  pages={540--557},
  year={2022},
  organization={Springer}
}

@inproceedings{ding2022decoupling,
  title={Decoupling zero-shot semantic segmentation},
  author={Ding, Jian and Xue, Nan and Xia, Gui-Song and Dai, Dengxin},
  booktitle={Proceedings of the IEEE/CVF conference on computer vision and pattern recognition},
  pages={11583--11592},
  year={2022}
}

@inproceedings{xu2022simple,
  title={A simple baseline for open-vocabulary semantic segmentation with pre-trained vision-language model},
  author={Xu, Mengde and Zhang, Zheng and Wei, Fangyun and Lin, Yutong and Cao, Yue and Hu, Han and Bai, Xiang},
  booktitle={European Conference on Computer Vision},
  pages={736--753},
  year={2022},
  organization={Springer}
}

@inproceedings{liang2023open,
  title={Open-vocabulary semantic segmentation with mask-adapted clip},
  author={Liang, Feng and Wu, Bichen and Dai, Xiaoliang and Li, Kunpeng and Zhao, Yinan and Zhang, Hang and Zhang, Peizhao and Vajda, Peter and Marculescu, Diana},
  booktitle={Proceedings of the IEEE/CVF conference on computer vision and pattern recognition},
  pages={7061--7070},
  year={2023}
}

@inproceedings{zhou2023zegclip,
  title={Zegclip: Towards adapting clip for zero-shot semantic segmentation},
  author={Zhou, Ziqin and Lei, Yinjie and Zhang, Bowen and Liu, Lingqiao and Liu, Yifan},
  booktitle={Proceedings of the IEEE/CVF conference on computer vision and pattern recognition},
  pages={11175--11185},
  year={2023}
}

@inproceedings{xu2023side,
  title={Side adapter network for open-vocabulary semantic segmentation},
  author={Xu, Mengde and Zhang, Zheng and Wei, Fangyun and Hu, Han and Bai, Xiang},
  booktitle={Proceedings of the IEEE/CVF conference on computer vision and pattern recognition},
  pages={2945--2954},
  year={2023}
}

@inproceedings{xu2023open,
  title={Open-vocabulary panoptic segmentation with text-to-image diffusion models},
  author={Xu, Jiarui and Liu, Sifei and Vahdat, Arash and Byeon, Wonmin and Wang, Xiaolong and De Mello, Shalini},
  booktitle={Proceedings of the IEEE/CVF conference on computer vision and pattern recognition},
  pages={2955--2966},
  year={2023}
}

@inproceedings{shan2024open,
  title={Open-vocabulary semantic segmentation with image embedding balancing},
  author={Shan, Xiangheng and Wu, Dongyue and Zhu, Guilin and Shao, Yuanjie and Sang, Nong and Gao, Changxin},
  booktitle={Proceedings of the IEEE/CVF Conference on Computer Vision and Pattern Recognition},
  pages={28412--28421},
  year={2024}
}

@inproceedings{xie2024sed,
  title={Sed: A simple encoder-decoder for open-vocabulary semantic segmentation},
  author={Xie, Bin and Cao, Jiale and Xie, Jin and Khan, Fahad Shahbaz and Pang, Yanwei},
  booktitle={Proceedings of the IEEE/CVF conference on computer vision and pattern recognition},
  pages={3426--3436},
  year={2024}
}

@inproceedings{zhao2025dpseg,
  title={DPSeg: Dual-Prompt Cost Volume Learning for Open-Vocabulary Semantic Segmentation},
  author={Zhao, Ziyu and Li, Xiaoguang and Shi, Lingjia and Imanpour, Nasrin and Wang, Song},
  booktitle={Proceedings of the Computer Vision and Pattern Recognition Conference},
  pages={25346--25356},
  year={2025}
}

@inproceedings{radford2021learning,
  title={Learning transferable visual models from natural language supervision},
  author={Radford, Alec and Kim, Jong Wook and Hallacy, Chris and Ramesh, Aditya and Goh, Gabriel and Agarwal, Sandhini and Sastry, Girish and Askell, Amanda and Mishkin, Pamela and Clark, Jack and others},
  booktitle={International conference on machine learning},
  pages={8748--8763},
  year={2021},
  organization={PmLR}
}

@inproceedings{jia2021scaling,
  title={Scaling up visual and vision-language representation learning with noisy text supervision},
  author={Jia, Chao and Yang, Yinfei and Xia, Ye and Chen, Yi-Ting and Parekh, Zarana and Pham, Hieu and Le, Quoc and Sung, Yun-Hsuan and Li, Zhen and Duerig, Tom},
  booktitle={International conference on machine learning},
  pages={4904--4916},
  year={2021},
  organization={PMLR}
}

@inproceedings{lee2025effective,
  title={Effective SAM Combination for Open-Vocabulary Semantic Segmentation},
  author={Lee, Minhyeok and Cho, Suhwan and Lee, Jungho and Yang, Sunghun and Choi, Heeseung and Kim, Ig-Jae and Lee, Sangyoun},
  booktitle={Proceedings of the Computer Vision and Pattern Recognition Conference},
  pages={26081--26090},
  year={2025}
}

@inproceedings{choi2024salience,
  title={Salience-based adaptive masking: revisiting token dynamics for enhanced pre-training},
  author={Choi, Hyesong and Park, Hyejin and Yi, Kwang Moo and Cha, Sungmin and Min, Dongbo},
  booktitle={European Conference on Computer Vision},
  pages={343--359},
  year={2024},
  organization={Springer}
}

@inproceedings{luo2024emergent,
  title={Emergent open-vocabulary semantic segmentation from off-the-shelf vision-language models},
  author={Luo, Jiayun and Khandelwal, Siddhesh and Sigal, Leonid and Li, Boyang},
  booktitle={Proceedings of the IEEE/CVF Conference on Computer Vision and Pattern Recognition},
  pages={4029--4040},
  year={2024}
}

@inproceedings{cheng2022masked,
  title={Masked-attention mask transformer for universal image segmentation},
  author={Cheng, Bowen and Misra, Ishan and Schwing, Alexander G and Kirillov, Alexander and Girdhar, Rohit},
  booktitle={Proceedings of the IEEE/CVF conference on computer vision and pattern recognition},
  pages={1290--1299},
  year={2022}
}

@inproceedings{selvaraju2017grad,
  title={Grad-cam: Visual explanations from deep networks via gradient-based localization},
  author={Selvaraju, Ramprasaath R and Cogswell, Michael and Das, Abhishek and Vedantam, Ramakrishna and Parikh, Devi and Batra, Dhruv},
  booktitle={Proceedings of the IEEE international conference on computer vision},
  pages={618--626},
  year={2017}
}

@article{barsellotti2024talking,
  title={Talking to DINO: Bridging self-supervised vision backbones with language for open-vocabulary segmentation},
  author={Barsellotti, Luca and Bianchi, Lorenzo and Messina, Nicola and Carrara, Fabio and Cornia, Marcella and Baraldi, Lorenzo and Falchi, Fabrizio and Cucchiara, Rita},
  journal={arXiv preprint arXiv:2411.19331},
  year={2024}
}

@inproceedings{wang2025iterprime,
  title={Iterprime: Zero-shot referring image segmentation with iterative grad-cam refinement and primary word emphasis},
  author={Wang, Yuji and Ni, Jingchen and Liu, Yong and Yuan, Chun and Tang, Yansong},
  booktitle={Proceedings of the AAAI Conference on Artificial Intelligence},
  volume={39},
  number={8},
  pages={8159--8168},
  year={2025}
}

@article{li2021align,
  title={Align before fuse: Vision and language representation learning with momentum distillation},
  author={Li, Junnan and Selvaraju, Ramprasaath and Gotmare, Akhilesh and Joty, Shafiq and Xiong, Caiming and Hoi, Steven Chu Hong},
  journal={Advances in neural information processing systems},
  volume={34},
  pages={9694--9705},
  year={2021}
}

@inproceedings{liu2021swin,
  title={Swin transformer: Hierarchical vision transformer using shifted windows},
  author={Liu, Ze and Lin, Yutong and Cao, Yue and Hu, Han and Wei, Yixuan and Zhang, Zheng and Lin, Stephen and Guo, Baining},
  booktitle={Proceedings of the IEEE/CVF international conference on computer vision},
  pages={10012--10022},
  year={2021}
}

@inproceedings{liu2024open,
  title={Open-vocabulary segmentation with semantic-assisted calibration},
  author={Liu, Yong and Bai, Sule and Li, Guanbin and Wang, Yitong and Tang, Yansong},
  booktitle={Proceedings of the IEEE/CVF Conference on Computer Vision and Pattern Recognition},
  pages={3491--3500},
  year={2024}
}

@inproceedings{li2024learning,
  title={Learning background prompts to discover implicit knowledge for open vocabulary object detection},
  author={Li, Jiaming and Zhang, Jiacheng and Li, Jichang and Li, Ge and Liu, Si and Lin, Liang and Li, Guanbin},
  booktitle={Proceedings of the IEEE/CVF Conference on Computer Vision and Pattern Recognition},
  pages={16678--16687},
  year={2024}
}

@inproceedings{li2022panoptic,
  title={Panoptic segformer: Delving deeper into panoptic segmentation with transformers},
  author={Li, Zhiqi and Wang, Wenhai and Xie, Enze and Yu, Zhiding and Anandkumar, Anima and Alvarez, Jose M and Luo, Ping and Lu, Tong},
  booktitle={Proceedings of the IEEE/CVF conference on computer vision and pattern recognition},
  pages={1280--1289},
  year={2022}
}

@inproceedings{simeoni2023unsupervised,
  title={Unsupervised object localization: Observing the background to discover objects},
  author={Sim{\'e}oni, Oriane and Sekkat, Chlo{\'e} and Puy, Gilles and Vobeck{\`y}, Anton{\'\i}n and Zablocki, {\'E}loi and P{\'e}rez, Patrick},
  booktitle={Proceedings of the IEEE/CVF conference on computer vision and pattern recognition},
  pages={3176--3186},
  year={2023}
}

@article{dosovitskiy2020image,
  title={An image is worth 16x16 words: Transformers for image recognition at scale},
  author={Dosovitskiy, Alexey and Beyer, Lucas and Kolesnikov, Alexander and Weissenborn, Dirk and Zhai, Xiaohua and Unterthiner, Thomas and Dehghani, Mostafa and Minderer, Matthias and Heigold, Georg and Gelly, Sylvain and others},
  journal={arXiv preprint arXiv:2010.11929},
  year={2020}
}

@misc{wu2019detectron2,
  author =       {Yuxin Wu and Alexander Kirillov and Francisco Massa and
                  Wan-Yen Lo and Ross Girshick},
  title =        {Detectron2},
  howpublished = {\url{https://github.com/facebookresearch/detectron2}},
  year =         {2019}
}

@article{loshchilov2017decoupled,
  title={Decoupled weight decay regularization},
  author={Loshchilov, Ilya and Hutter, Frank},
  journal={arXiv preprint arXiv:1711.05101},
  year={2017}
}

@article{paszke2019pytorch,
  title={Pytorch: An imperative style, high-performance deep learning library},
  author={Paszke, A},
  journal={arXiv preprint arXiv:1912.01703},
  year={2019}
}

@inproceedings{you2025focus,
  title={FOCUS: Towards universal foreground segmentation},
  author={You, Zuyao and Kong, Lingyu and Meng, Lingchen and Wu, Zuxuan},
  booktitle={Proceedings of the AAAI Conference on Artificial Intelligence},
  volume={39},
  number={9},
  pages={9580--9588},
  year={2025}
}

@inproceedings{lan2024clearclip,
  title={Clearclip: Decomposing clip representations for dense vision-language inference},
  author={Lan, Mengcheng and Chen, Chaofeng and Ke, Yiping and Wang, Xinjiang and Feng, Litong and Zhang, Wayne},
  booktitle={European Conference on Computer Vision},
  pages={143--160},
  year={2024},
  organization={Springer}
}

@article{hu2025pranet,
  title={PraNet-V2: Dual-Supervised Reverse Attention for Medical Image Segmentation},
  author={Hu, Bo-Cheng and Ji, Ge-Peng and Shao, Dian and Fan, Deng-Ping},
  journal={arXiv preprint arXiv:2504.10986},
  year={2025}
}

@article{achiam2023gpt,
  title={Gpt-4 technical report},
  author={Achiam, Josh and Adler, Steven and Agarwal, Sandhini and Ahmad, Lama and Akkaya, Ilge and Aleman, Florencia Leoni and Almeida, Diogo and Altenschmidt, Janko and Altman, Sam and Anadkat, Shyamal and others},
  journal={arXiv preprint arXiv:2303.08774},
  year={2023}
}

@article{gotmare2018closer,
  title={A closer look at deep learning heuristics: Learning rate restarts, warmup and distillation},
  author={Gotmare, Akhilesh and Keskar, Nitish Shirish and Xiong, Caiming and Socher, Richard},
  journal={arXiv preprint arXiv:1810.13243},
  year={2018}
}
\bibliographystyle{icml2026}

%%%%%%%%%%%%%%%%%%%%%%%%%%%%%%%%%%%%%%%%%%%%%%%%%%%%%%%%%%%%%%%%%%%%%%%
% APPENDIX
%%%%%%%%%%%%%%%%%%%%%%%%%%%%%%%%%%%%%%%%%%%%%%%%%%%%%%%%%%%%%%%%%%%%%%%
\newpage
\appendix
\onecolumn
%%%%%%%%%%%%%%%%%%%%%%%%%%%%%%%%%%%%%%%%%%%%%%%%%%%%%%%%%%%%
\newpage
\appendix
\section*{Appendix Overview}
In this Appendix, we provide additional details of the paper, including other inference-time saliency details (Section \ref{sec:saliencydetail}), other model details (Section \ref{sec:modeldetail}), other implementation details (Section \ref{sec:impledetail}), additional ablation study (Section \ref{sec:addab}), additional qualitative results (Section \ref{sec:addvis}), and limitations of proposed method (Section \ref{sec:limitation}). \par

\section{Other Inference-Time Saliency Details} \label{sec:saliencydetail}
\noindent \textbf{Saliency generation during inference.} The Image-Text Matching predicts whether a pair of image and class label is positive (matched) or negative (not matched) \cite{luo2024emergent}. This is equivalent to asking: \textit{which attention embeddings contribute the most to decide if the image-text pair is matching?} We use a MLP to regress $\boldsymbol{\hat{y}}^{itm}_{(v, t)}$ in Eq. \ref{eq:ITM} based on the attention maps $A$ as the joint representation of the image-text pair, predicting a two-class probability (matched or not matched). The ITM loss and segmentation supervision are used only during training, while inference relies solely on image-text alignment. At inference, our model uses the gradients of the predicted matching scores, not gradients of the loss. Specifically, we treat the matching score as a scalar objective and back-propagate from it following the standard practice in gradient-based attribution methods \cite{selvaraju2017grad}. Note that DiSa doesn't need ground truth segmentation mask, class label, or pixel-level supervision during inference. \par

\noindent \textbf{Generalization to novel classes.} Although the ITM loss is supervised using base classes' masks, it does not learn class-specific saliency patterns. Instead, it trains the model to identify class-agnostic foregroundness, i.e., visually salient structures (edges, textures, object boundaries) that are shared across categories. This supervision regularizes the saliency mechanism to highlight object-centric regions rather than memorizing base classes, which allows it to generalize naturally to novel open-vocabulary classes when combined with class-level correlation maps at inference. \par

\section{Other Model Details}\label{sec:modeldetail}
\subsection{Upsampling Decoder}
We adopt a lightweight upsampling decoder following the design in \citep{cho2024cat}. Specifically, we extract intermediate visual embeddings from the 4-th and 8-th layers \citep{dosovitskiy2020image} of the CLIP ViT-L/14 image encoder (or the 8-th and 16-th layers of CLIP ViT-L/14) for higher-resolution guidance. The decoder consists of two identical transposed convolution layers that progressively upsample the feature maps. It takes the correlation maps with a resolution of 24 $\times$ 24 as input, and outputs predictions at a resolution of 96 $\times$ 96. The effectiveness of this simple decoder stems from our saliency-aware disentanglement and hierarchical refinement, which models rich contexts while preserving accurate object boundaries, thereby enhancing feature extraction qualities. \par

\noindent \textbf{Ablation study details.} For class-level disentanglement ((II) in Table \ref{table:ablation}), we use ChatGPT 4o \citep{achiam2023gpt} to partition all classes into foreground and background classes with two complementary branches. Similar to our DiSa's design, each branch is processed independently to capture distinct class-specific cues, and the resulting representations are finally merged to form unified embeddings. \par

\section{Other Implementation Details}\label{sec:impledetail}
\noindent \textbf{Training details.} We use pre-trained CLIP ViT-B/16 \citep{radford2021learning} as our base-VLM and CLIP ViT-L/14 \citep{radford2021learning} as our large-VLM, following the same setting as in most recent SOTA models \citep{cho2024cat}. We train three attention layers for generating saliency maps following the empirical experience and ablation studies in PnP-OVSS \citep{luo2024emergent}. We adopt settings of 8 heads and dimension size $=512$ for these attention layers. For the Pixel-Wise Refinement and subsequent Swin Transformer \citep{liu2021swin} used for aggregation, we adopt a commonly used structure: one non-shifted window attention layer, followed by a shifted window attention layer. Some other training details include Warmup Cosine Learning Rate scheduler \citep{gotmare2018closer} and 1e-4 weight decay. \par

\noindent \textbf{Data preprocessing.} The data augmentation used in our work includes random cropping, and photometric distortion, following \citep{cheng2022masked}. During training, saturation, hue, and contrast are randomly adjusted for robustness. The training resolution is set to be 384 $\times$ 384. \par

\noindent \textbf{Text template.} We utilize the commonly used prompt template for text labels, which is "A photo of a {class}", without relying on cutting-edge templates. We do not incorporate any LLM-generated or handcrafted prompts in our work. \par

\begin{table}[t]
\centering
\resizebox{0.53\linewidth}{!}{%
\begin{tabular}{lcccccc}
\toprule
\textbf{$k$} & \textbf{A-847} & \textbf{PC-459} & \textbf{A-150} & \textbf{PC-59} & \textbf{PAS-20} & \textbf{PAS-20\textsuperscript{b}} \\
\midrule
(I) 16 & 11.4 & 18.7 & 32.6 & 57.8 & 96.3 & 78.0 \\
(II) 48 & \secondbest{12.0} & \secondbest{19.8} & \best{34.1} & \secondbest{58.3} & \best{97.1} & \secondbest{78.6} \\
\midrule
(III) \textbf{96} & \best{12.6} & \best{20.3} & \secondbest{33.7} & \best{59.3} & \secondbest{97.0} & \best{79.9} \\
\bottomrule
\end{tabular}
}
\caption{\textbf{Ablation study for foreground selection hyperparameter $k$.}}
\vspace{-1.2\baselineskip}
\label{table:ablationk}
\end{table}

\noindent \textbf{Evaluation metrics.} We use mean Intersection over Union (mIoU) to measure segmentation performance. For model efficiency analysis, we use parameter size and GFLOPs. \par

\section{Additional Ablation Study}\label{sec:addab}
\noindent \textbf{Design choices of foreground selection $k$.} To further validate the impact of foreground selection hyperparameter, we leverage different $k$ to evaluate our model's robustness in Table \ref{table:ablationk}: \textbf{(I)} $k$=16 (3\% ratio of all tokens), \textbf{(II)} $k$=48 (8\% ratio of all tokens), \textbf{(III)} $k$=96 (17\% ratio of all tokens). Overall, generally speaking, the performance gradually improves with increasing $k$, and it yields small but consistent gains when $k$=96, achieving a good balance between expressiveness (for small $k$) and noise (for large $k$). PC-459 and A-847 improve the most across $k$, suggesting they benefit significantly from foreground-background disentanglement. Results on PAS-20 are relatively robust, which reveals that our disentanglement design captures complex scenario understanding and mitigates \textbf{\textit{Foreground Bias}}. \par

\begin{table}[t]
\centering
\resizebox{0.6\linewidth}{!}{%
\begin{tabular}{lcccccc}
\toprule
 & \textbf{A-847} & \textbf{PC-459} & \textbf{A-150} & \textbf{PC-59} & \textbf{PAS-20} & \textbf{PAS-20\textsuperscript{b}} \\
\midrule
(I) Attn agg & \secondbest{12.5} & \best{20.3} & \secondbest{33.5} & \secondbest{59.0} & \best{97.3} & 79.4 \\
(II) Hard agg & 12.2 & \secondbest{19.9} & 32.9 & 58.7 & 96.4 & \best{80.1} \\
\midrule
(III) \textbf{Weighted agg} & \best{12.6} & \best{20.3} & \best{33.7} & \best{59.3} & \secondbest{97.0} & \secondbest{79.9} \\
\bottomrule
\end{tabular}
}
\caption{\textbf{Ablation study for foreground/background aggregation designs.}}
\vspace{-0.6\baselineskip}
\label{table:ablationagg}
\end{table}

\begin{figure*}[t]
\centering
\begin{overpic}[width=0.8\linewidth]{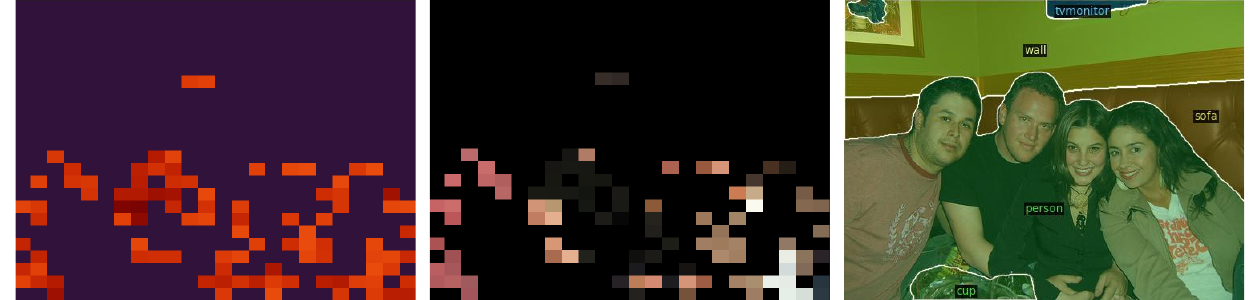}
  \put(6, -2.5){\small (a) Foreground saliency}       % left column caption
  \put(39, -2.5){\small (b) Foreground tokens}  % middle column
  \put(76, -2.5){\small (c) Prediction}  % right column
\end{overpic}
\vspace{0.6\baselineskip}
\caption{\textbf{Failure Case with imprecise foreground/background disentanglement.}}
\vspace{-1.2\baselineskip}
\label{fig:failure}
\end{figure*}

\noindent \textbf{Design choices of foreground/background feature aggregation.} We conduct additional experiments that compare (I) Attention-based aggregation, using learnable weights with attention mechanism; (II) Hard aggregation, directly aggregating foreground and background tokens (reorganization with mask indices), and (III) our weighted aggregation (by using learnable weights by gating mechanism). in Table \ref{table:ablationagg}. Our design achieves significant improvements, demonstrating our design's robustness. \par

\section{Additional Qualitative Results}\label{sec:addvis}
\begin{figure*}[!htbp]
\centering
\includegraphics[width=0.8\linewidth]{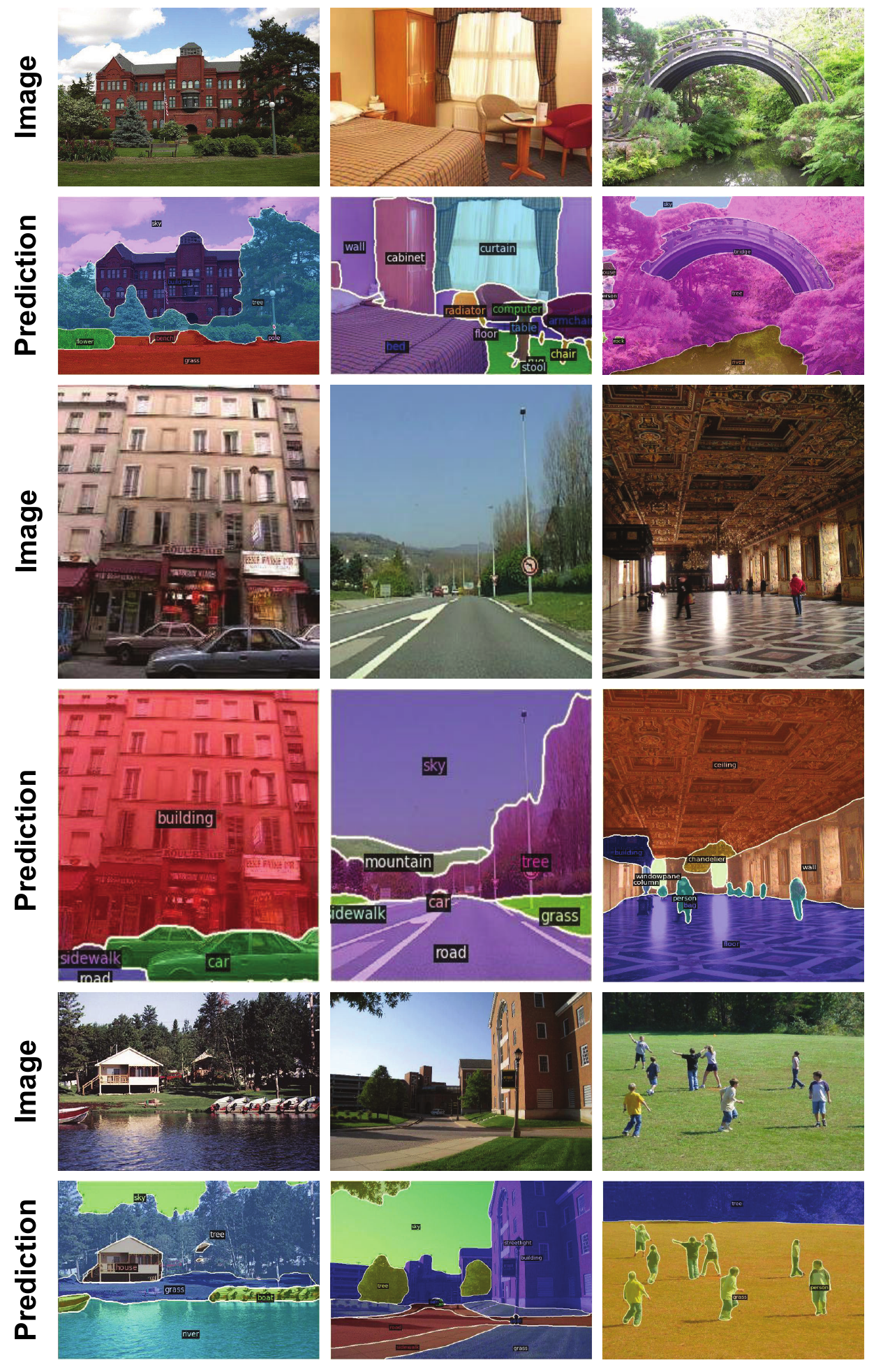}
\caption{\textbf{Qualitative results on ADE20K with 150 classes.}}
\label{fig:a150}
\end{figure*}

\begin{figure*}[!htbp]
\centering
\includegraphics[width=0.83\linewidth]{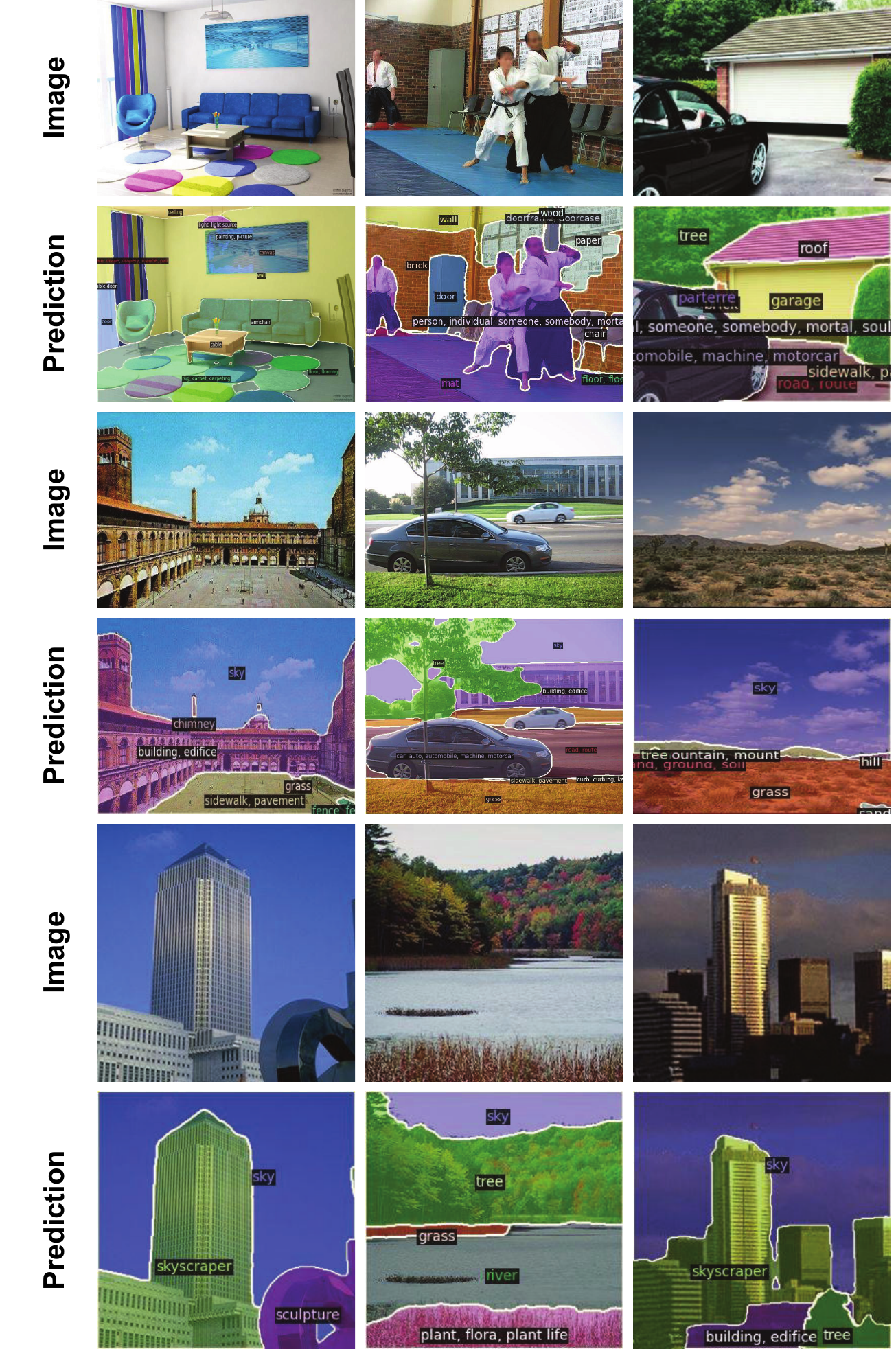}
\caption{\textbf{Qualitative results on ADE20K with 847 classes.}}
\label{fig:a847}
\end{figure*}

\begin{figure*}[!htbp]
\centering
\includegraphics[width=0.83\linewidth]{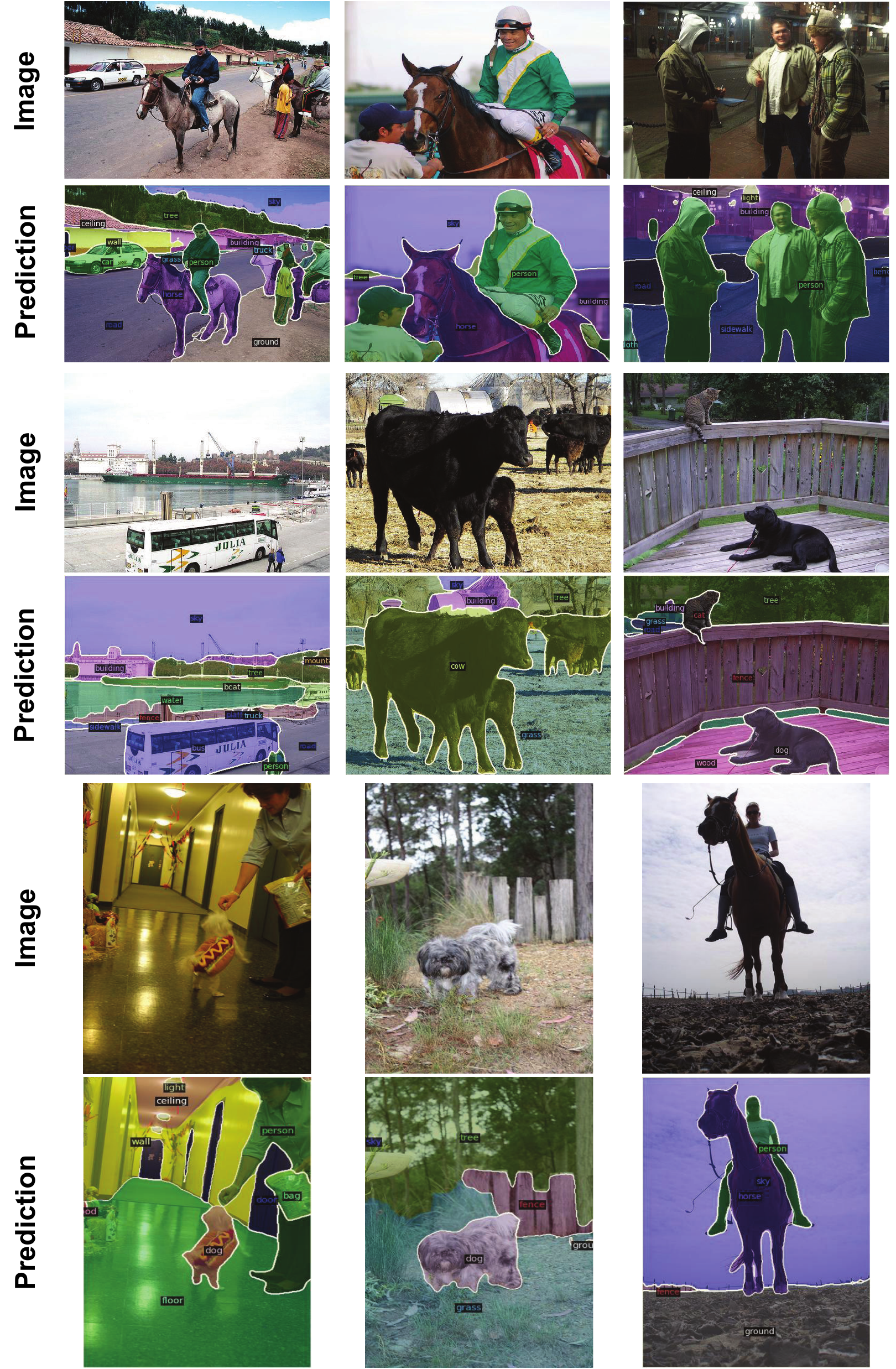}
\caption{\textbf{Qualitative results on PASCAL Context with 59 classes.}}
\label{fig:pc59}
\end{figure*}

\begin{figure*}[!htbp]
\centering
\includegraphics[width=0.83\linewidth]{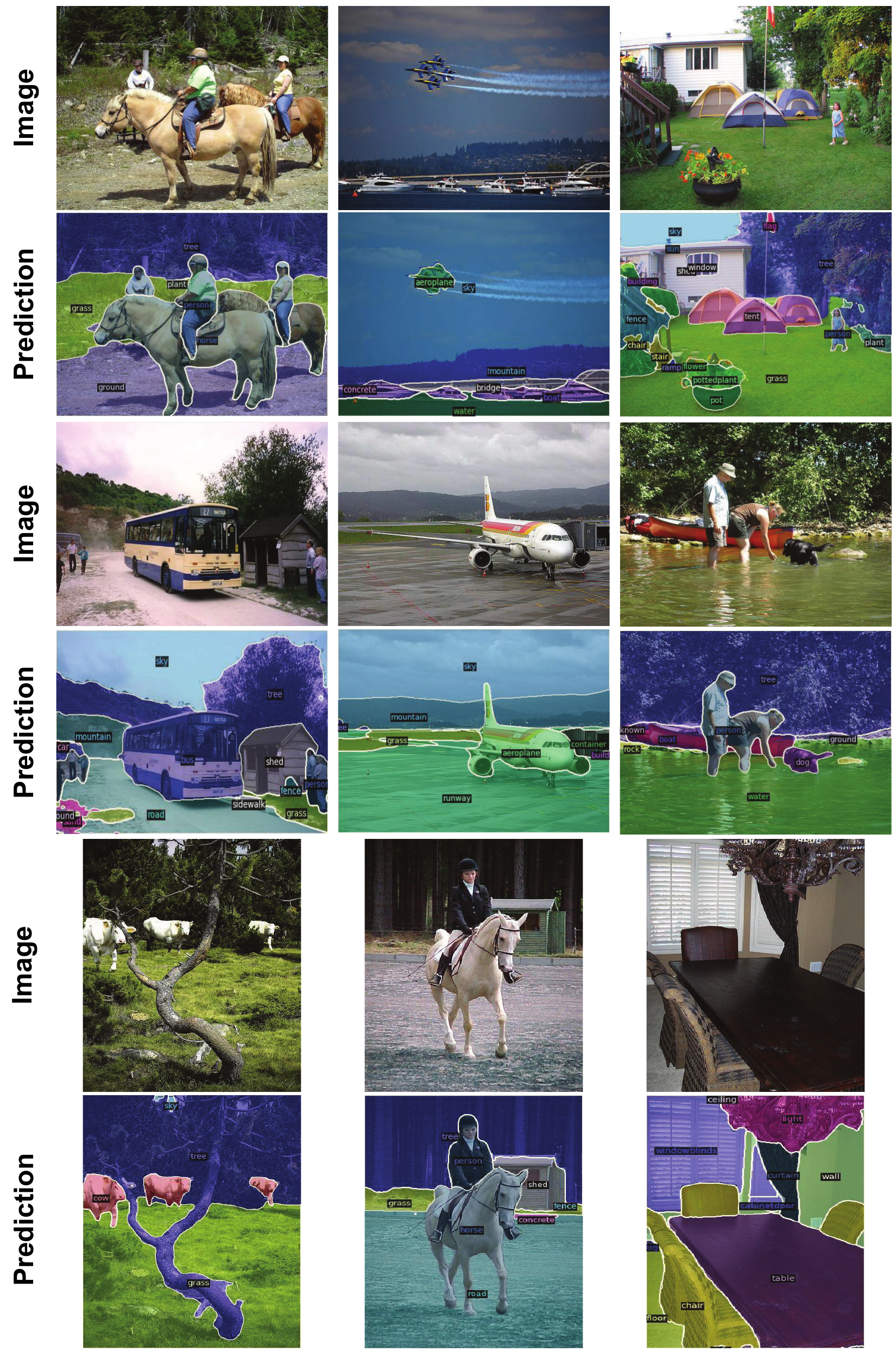}
\caption{\textbf{Qualitative results on PASCAL Context with 459 classes.}}
\label{fig:pc459}
\end{figure*}

To further validate our model, we present more visualizations of qualitative results on A-150 \citep{zhou2019semantic} in Fig. \ref{fig:a150}, A-847 \citep{zhou2019semantic} in Fig. \ref{fig:a847}, PC-59 \citep{everingham2010pascal} in Fig. \ref{fig:pc59}, and PC-459 \citep{everingham2010pascal} in Fig. \ref{fig:pc459}. DiSa consistently produces accurate and robust predictions in complex scenarios, demonstrating its efficacy. \par

\begin{figure*}[!htbp]
\centering
\begin{overpic}[width=0.66\linewidth]{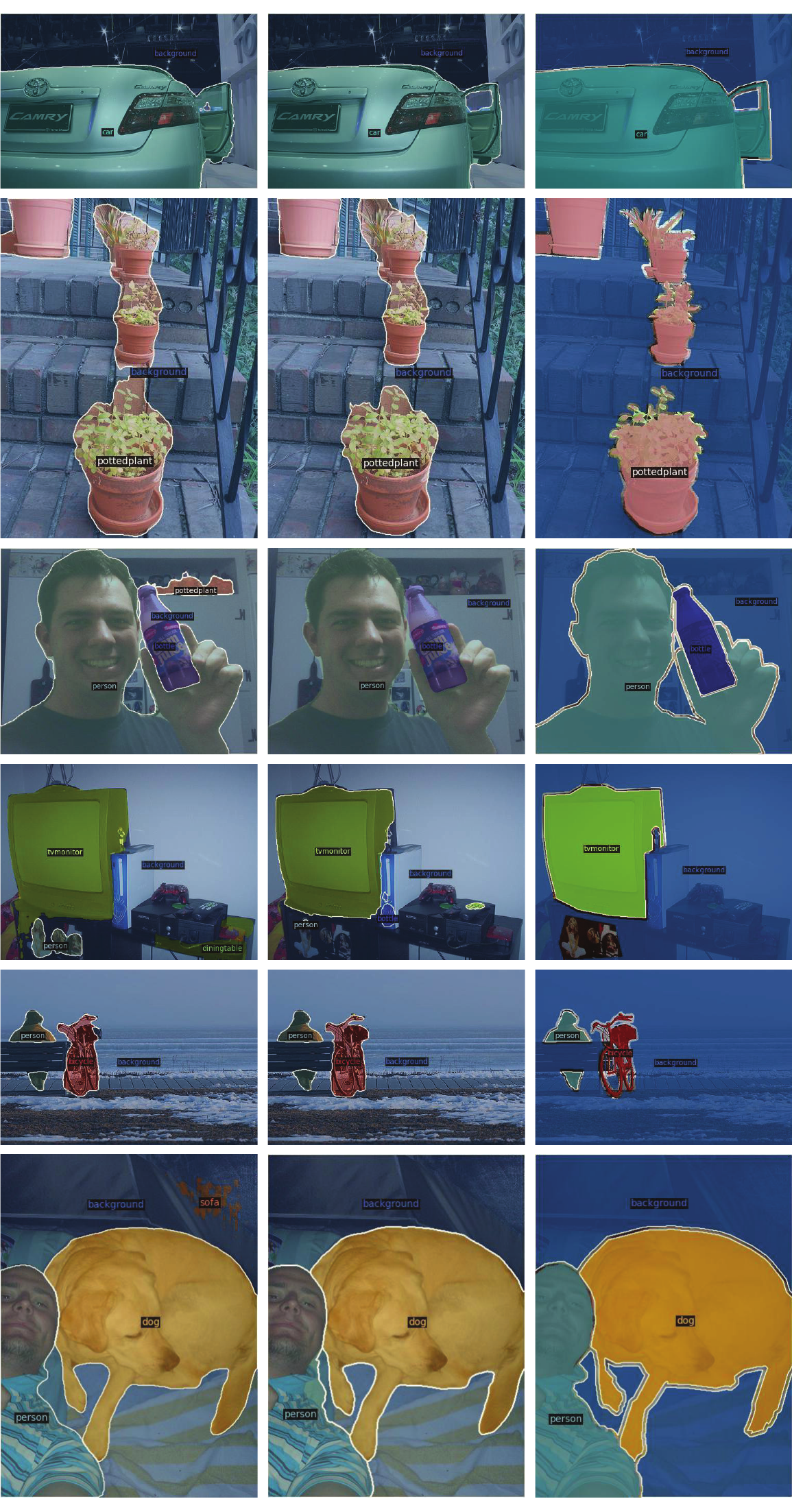}
  \put(5, -1){\small (a) \textbf{CAT-Seg}}
  \put(23.5, -1){\small (b) \textbf{Ours}}
  \put(41.5, -1){\small (c) \textbf{GT}}
\end{overpic}
\vspace{3mm}
\caption{\textbf{Comparison of Qualitative results on PAS-20$^{b}$.} We compare DiSa with CAT-Seg.}
\label{fig:pas20b}
\end{figure*}

We additionally present the comparison of qualitative results on PAS-20$^{b}$ \citep{ghiasi2022scaling} between DiSa and one previous SOTA approach, CAT-Seg \citep{cho2024cat}, in Fig. \ref{fig:pas20b}. Note that PAS-20$^{b}$ has one extra “background” class, and the results clearly illustrate our identified limitations and motivations. Specifically, CAT-Seg struggles to (i) separate foreground and background areas (e.g., window in row 1, potted plant in row 2, and sofa in row 5), and (ii) define accurate boundaries between objects (e.g., TV monitor in row 3 and bicycle in row 4). These two limitations correspond to the \textbf{\textit{Foreground Bias}} and \textbf{\textit{Limited Spatial Localization}} inherent in VLMs, respectively. In contrast, DiSa improves foreground-background contexts and generates more precise object boundaries, demonstrating DiSa's ability to tackle challenging scenarios and mitigate the aforementioned limitations. \par

\section{Failure Cases}\label{sec:failure}
We additionally provide a failure case of imperfect foreground/background separation in Fig. \ref{fig:failure}. In this crowded scene, some salient regions of the class "person" are not assigned as the foreground. It demonstrates that, for objects that vary widely in size, the disentanglement might become unstable, leading to inaccurate or ambiguous foreground/background separation. However, the ensemble nature of our dual branches provides robustness by preserving complementary cues in the alternative branch compensate for such errors, leading to more reliable fused predictions. \par

\section{Limitation}\label{sec:limitation}
Our foreground selection hyperparameter $k$ is fixed and we plan to make it adaptive. Instead of selecting a constant number of foreground tokens for all images, we will predict a dynamic $k$ conditioned on the input image, e.g., using saliency/uncertainty statistics or a lightweight gating module, to reflect varying object scales and scene complexity. This input-dependent token allocation can better balance foreground–background evidence, reduce sensitivity to hyperparameters, and further improve robustness across diverse open-vocabulary scenes. \par

\clearpage

\end{document}